%% file: main.tex
\documentclass[preprint,12pt,authoryear]{elsarticle}

\usepackage{amssymb}
\usepackage{amsmath}

\usepackage[table,xcdraw]{xcolor}
\usepackage{colortbl}
\usepackage{textcomp}

\usepackage{comment}
\usepackage{color}
\usepackage{graphicx}
\usepackage{svg}  
\usepackage{tikz}
\usepackage{afterpage}
\usepackage{booktabs}
\usepackage{float}
\usepackage{siunitx}
\usepackage{natbib}

\usepackage{subcaption}

\usetikzlibrary{positioning}

\usepackage[colorlinks=true, linkcolor=blue, citecolor=blue, urlcolor=blue]{hyperref}

\usepackage[linesnumbered,ruled,vlined]{algorithm2e}

\newcommand{\eric}[1]{\textcolor{black}{#1}}

\usepackage{lineno}

\makeatletter
\def\ps@pprintTitle{%
  \let\@oddhead\@empty
  \let\@evenhead\@empty
  \def\@oddfoot{\reset@font\hfil\thepage\hfil}
  \let\@evenfoot\@oddfoot
}
\makeatother

\begin{document}

\begin{frontmatter}



\title{Physics-Informed Neural Networks for Control of Single-Phase Flow Systems Governed by Partial Differential Equations} 






 \author[NTNU,UFSC]{Luis Kin Miyatake}
 \ead{luiskm@stud.ntnu.no}
 \affiliation[NTNU]{organization={Norwegian University of Science and Technology},
             addressline={Petroleumsteknisk senter, 512, Valgrinda, S.P. Andersensveg 15a},
             city={Trondheim},
             country={Norway}}

 \affiliation[UFSC]{organization={Department of Automation and Systems Engineering, Federal University of Santa Catarina},
            addressline={Cx.P 476},
             city={Florianópolis},
             postcode={88040-900},
             state={SC},
             country={Brazil}}

\author[UFSC]{Eduardo Camponogara} 
 \ead{eduardo.camponogara@ufsc.br}
\author[UFSC]{Eric Aislan Antonelo} 
 \ead{eric.antonelo@ufsc.br}

\author[NTNU]{Alexey Pavlov} 
\ead{alexey.pavlov@ntnu.no}

\begin{footnotesize}

\begin{abstract}
The modeling and control of single-phase flow systems governed by Partial Differential Equations (PDEs) present challenges, especially under transient conditions. In this work, we extend the Physics-Informed Neural Nets for Control (PINC) framework, \eric{originally proposed for modeling and control of Ordinary Differential Equations (ODEs) without the need of any labeled data, to the PDE case, particularly} to single-phase incompressible and compressible flows, integrating neural networks with physical conservation laws. 
The PINC model \eric{for PDEs} is structured into two stages: a steady-state network, which learns equilibrium solutions for a wide range of control inputs, and a transient network, which captures dynamic responses under time-varying boundary conditions. 
  We propose a simplifying assumption that reduces the dimensionality of the spatial coordinate regarding the initial condition, allowing the efficient training of the PINC network.
This simplification enables the derivation of optimal control policies using Model Predictive Control (MPC). 
   We validate our approach through numerical experiments, demonstrating that the PINC model, which is trained exclusively using physical laws, i.e., without labeled data, accurately represents flow dynamics and enables real-time control applications. 
The results highlight the PINC’s capability to efficiently approximate PDE solutions without requiring iterative solvers, making it a promising alternative for fluid flow monitoring and optimization in engineering applications.
\end{abstract}
\end{footnotesize}


\end{frontmatter}




\input{1-introduction}
\input{2-problem}

\input{4-methods}
\input{5-results}

\input{6-conclusion}



\afterpage{\clearpage

\input{main.bbl}
}









\end{document}

%% file: 1-introduction.tex
\section{Introduction}\label{sec:intro}
The modeling and control of fluid dynamics in complex systems, such as oil and gas pipelines, present significant challenges due to the intricate interactions between mass and momentum conservation laws, especially under transient conditions \citep{Malalasekera1995, shoham2006mechanistic}. In practical applications, real-time monitoring and control of these systems are crucial for maintaining operational efficiency and safety. However, rigorous simulation models, such as those based on transient simulators like OLGA \citep{OLGA}, while accurate, are often computationally intensive and unsuitable for real-time applications. Additionally, simulation models are typically black-box functions, meaning they do not provide access to derivative calculations. This is a critical limitation when applying optimization and control strategies that require gradient information. This further underscores the need for an efficient surrogate model that accurately represents the physical dynamics of such systems, is fast to compute, allows for direct computation without iterations, and is smooth with derivatives that support control and optimization applications.


To address this challenge, we propose an extension of a Physics-Informed Neural Control (PINC) \citep{ANTONELO2024127419} framework for PDEs, which integrates neural networks with governing physics equations, such as mass and momentum conservation, to model single-phase incompressible and compressible flow systems subject to time-varying boundary conditions, which act as control signals. In the PINC neural network, additional features representing controls and initial states are considered inputs beyond the usual features of a standard PINN for PDEs (position and time). 

The standard PINN belongs to a class of deep learning models designed to directly incorporate the underlying physical laws governing a system, such as conservation laws or differential equations, into the training process. Unlike traditional machine learning models that rely solely on data, PINNs leverage Partial Differential Equations (PDEs) to embed physical knowledge into the neural network. This approach allows PINNs to solve forward and inverse problems in physics-based systems more effectively, even with sparse or noisy data \citep{RAISSI2019686}. PINNs have proven to be highly effective in modeling complex systems, from fluid dynamics to electromagnetics, by considering the governing equations in the loss function, ensuring the solution's physical consistency \citep{karniadakis2021physics}.

The PINC network enables modeling a time-varying control system where the control signal remains constant within a specified time window but varies across different windows. The dynamic system evolves over time through a sequence of these time windows, allowing the derivation of an optimal control policy for the flow system using Model Predictive Control (MPC) \citep{camacho2007mpc, rawlings2009mpc}.

Our methodology progresses sequentially in terms of complexity. We begin with the steady-state regime, where the \textit{PINC Steady-State network} is trained to accurately capture the steady-state solution of the PDE system, using a wide range of control signals as inputs, which parametrize the network to accommodate a family of boundary conditions (or control signals). 

We then increase the model's complexity by incorporating transient effects, leading to the development of the \textit{PINC Transient network}. To handle the added complexity of the transient regime, we make a simplifying assumption: the initial condition for each current time window is set to the steady-state solution obtained using the control signal from the previous window. In this process, the PINC Steady-State model provides the PINC Transient with the necessary initial conditions during training. This assumption effectively reduces the input space of the PINC Transient model, making it easier to manage and improve training efficiency, while ensuring that the initial conditions represent the system's behavior realistically.

The PINC Transient model serves as the foundation for generating optimal control sequences in the Model Predictive Control (MPC) framework. This model guides the system toward the desired target while ensuring smooth control transitions and adherence to dynamic constraints.

The main contributions of this work are:
\begin{itemize}

    \item \textit{Extension of the basic PINC methodology, initially developed for ODEs, to PDEs, enabling forward simulations of PDEs over 
    \eric{arbitrarily long time horizons} and without the need for retraining.} 
    \eric{This extension for PDEs is achieved by sequentially training two PINC networks: the PINC Steady-State is first trained, serving as an auxiliary network for generating the target values for the initial condition during the training of the main PINC (Transient).} \eric{In particular, the proposed methodology is applied to a} PDE system representing a single-phase flow governed by mass and momentum conservation equations for incompressible and compressible systems.

    \item \textit{A simplifying assumption in the PINC architecture to make training more efficient.}
    It consists of using the control signal from the previous time window as input instead of the initial state, because the latter usually has a very high dimensionality associated with the spatial coordinate of PDEs.
       We assume the system reaches a steady-state regime when a constant control signal is applied within a time window. Therefore, the control signal from the previous time window is sufficient to define the initial condition for the current window. This approach significantly reduces input dimensionality, enhancing the efficiency of PINC training for PDEs.

    \item \textit{\eric{No error accumulation during PINC inference on long-term simulation of PDEs.}}
    As the PINC outputs depend only on the control signals from the previous and current time windows, \eric{due to our simplifying assumption,}
    errors do not propagate \eric{during forward simulation}, making 
    long-term simulations \eric{more precise}   \eric{when compared to the original PINC in \cite{ANTONELO2024127419}, which needs to feedback the output predictions in an autoregressive way.}
   
    \item
    \textit{\eric{Real-time MPC control of PDEs using the proposed PINC methodology.}}
    The pre-trained PINC model for PDEs, capable of accurately predicting system behavior across a wide range of control signals for long-term simulations, is integrated into an MPC controller. This integration enables real-time control of PDE-governed systems by leveraging the computational efficiency and differentiability of PINNs, which is demonstrated through an example application, where an optimal sequence of control signals is derived while respecting operational constraints in a highly nonlinear system governed by complex fluid dynamics equations.

\end{itemize}

\color{black}

%% file: 2-problem.tex
\section{Problem Statement} \label{sec:problem_statement}

This work addresses the conservation equations of mass and momentum for one-dimensional flow systems \citep{Malalasekera1995, OLGA, Aarsnes2014}, providing a detailed derivation for both incompressible and compressible single-phase flow scenarios. The conservation of mass ensures that the flow system adheres to the principle of mass continuity, while the momentum equation accounts for the forces acting on the fluid, including pressure gradients, frictional effects, and inertial contributions. 

For incompressible flow, the governing equations are simplified by assuming constant fluid density, which is representative of many liquid flow systems. On the other hand, for compressible flow, density variations with pressure are incorporated, making the system suitable for analyzing gas dynamics and other flow scenarios where compressibility effects are significant.

The subsequent sections detail the mathematical formulations of these conservation laws, along with the assumptions and simplifications specific to each flow regime. These derivations serve as a foundation for understanding the dynamics of one-dimensional flow systems and their applications in engineering. 

\subsection{General Modeling of Governing Equations}\label{subsection:general_modeling}

The governing equations for fluid flow, namely the conservation of mass and momentum equations, can be expressed in both conservative and non-conservative forms. 
    In the conservative form, the equations are written in terms of fluxes of conserved quantities. For example, the mass conservation equation (continuity equation) in one spatial dimension can be expressed as:
    \begin{equation}
    \frac{\partial \rho}{\partial t} + \frac{\partial (\rho V)}{\partial x} = 0,
    \label{eq:mass}
\end{equation}
where \(\rho\) is the fluid density and \(V\) is the velocity field. Similarly, the momentum conservation equation can be written as:
\begin{equation}
    \frac{\partial (\rho V)}{\partial t} + \frac{\partial (\rho V^2 + P)}{\partial x} = -\rho g \sin\theta - \frac{1}{2} \rho f \frac{|V|V}{D},
    \label{eq:momentum}
\end{equation}

In this equation, \(D\) is the diameter of the pipe, influencing frictional losses and flow dynamics. The angle \(\theta\) represents the pipe's inclination relative to the horizontal, with \(-\rho g \sin\theta\) accounting for the gravitational force component along the flow. The friction factor \(f\) quantifies flow resistance due to viscous effects and surface roughness, depending on the Reynolds number and, for rough pipes, the relative roughness \(\varepsilon/D\), where \(\varepsilon\) represents the absolute roughness of the pipe surface.

In the conservative form, the equations ensure strict conservation of mass and momentum by directly balancing fluxes across control volumes \citep{ferziger2002, white2006}. On the other hand, the non-conservative form expresses the equations in terms of the primitive variables (\textit{e.g.}, density \(\rho\), velocity \(V\), and pressure \(P\)) and their derivatives. For instance, in the non-conservative form, the momentum governing equation can be stated as:
\begin{equation}
    \rho \frac{\partial V}{\partial t} + \rho V \frac{\partial V}{\partial x} + \frac{\partial P}{\partial x} = -\rho g \sin\theta - \frac{1}{2} \rho f \frac{|V|V}{D}
    \label{eq:momentum_nc}
\end{equation}

\begin{figure}[h!]
    \centering
    \includegraphics[width=1.2\textwidth, trim=80 120 60 50, clip]{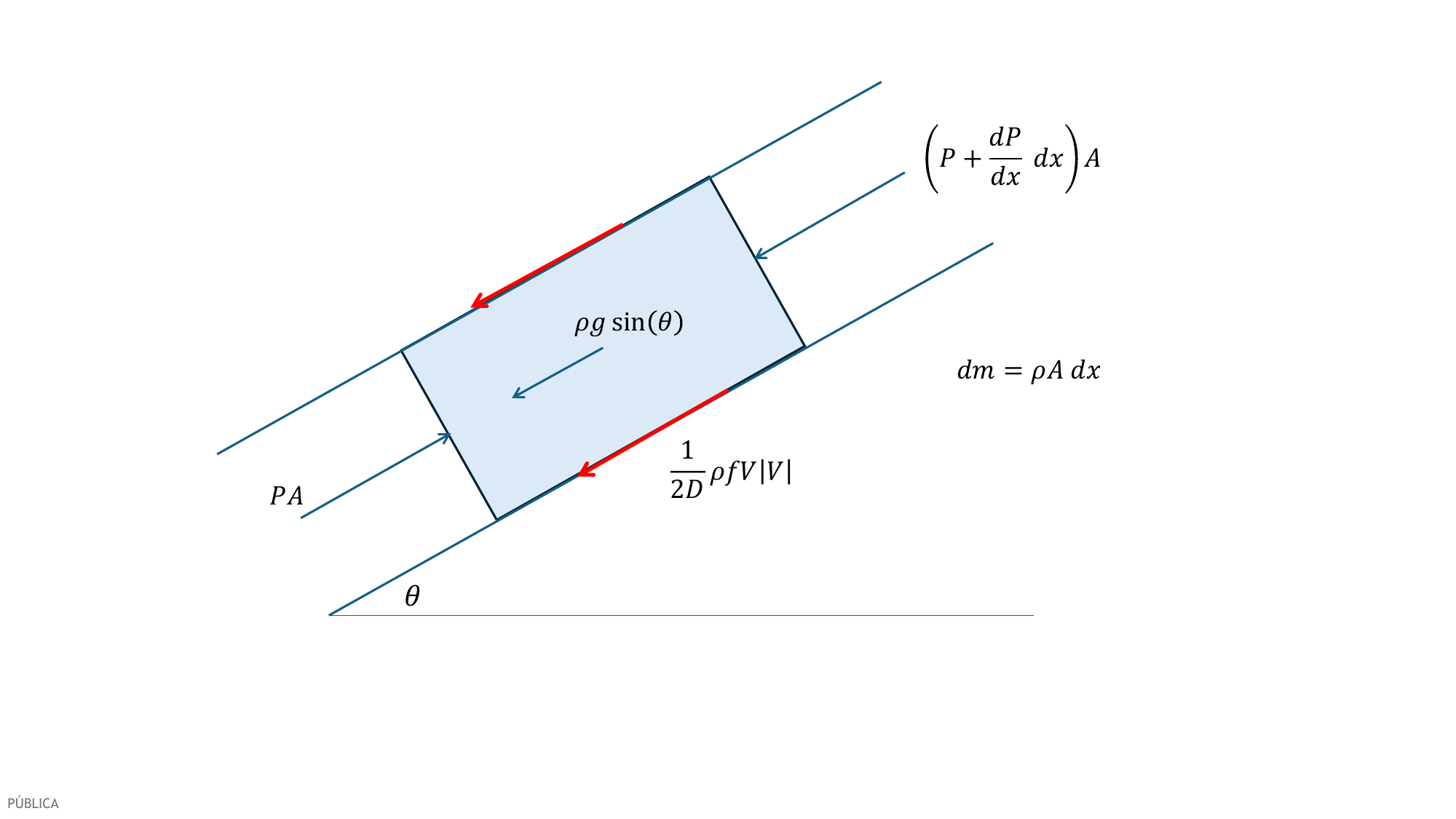}
    \caption{Representation of the force balance in the non-conservative formulation. The diagram illustrates the interaction between inertial forces, pressure gradients, and frictional effects.}
    \label{fig:force_balance}
\end{figure}

Here, the terms do not explicitly represent conserved fluxes. Instead, the non-conservative formulation directly represents a balance of forces (Figure \ref{fig:force_balance}), capturing the interplay between inertial, pressure, and frictional effects. This form emphasizes the physical mechanisms driving the flow.

Although these equations still describe the same physics, the non-conservative form is often more susceptible to numerical inaccuracies, especially across discontinuities such as shock waves, because it does not guarantee the strict conservation of mass and momentum at the discrete level \citep{leveque2002, toro2013}.

In computational fluid dynamics (CFD), the conservative form is generally preferred for simulating flows with strong discontinuities, shocks, or interfaces between fluids with significantly different properties. Conservative schemes ensure that any discontinuities are captured more accurately and that the integral form of the conservation laws is satisfied. Non-conservative forms, while sometimes simpler to derive and manipulate algebraically, can lead to spurious oscillations and non-physical results \citep{anderson1995, leveque2002}. 

For turbulent flow, the friction factor \( f \) can be approximated using several empirical correlations. In the laminar flow regime, the friction factor is given by the well-known relationship \( f = 64/Re \), where \( Re \) is the Reynolds number (\ref{eq:reynolds}). For turbulent flow in smooth pipes, the Blasius equation is a widely used correlation \citep{blasius1913}:
\begin{equation}
    f = \frac{0.316}{Re^{0.25}}.
    \label{eq:blasius}
\end{equation}

The Reynolds number \( Re \) is a dimensionless quantity defined as:
\begin{equation}\label{eq:reynolds}
    Re = \frac{\rho V D}{\mu},
\end{equation}
where \(\mu\) is the dynamic viscosity of the fluid.

For rough pipes, the friction factor depends on the relative roughness \(\varepsilon/D\) and the pipe diameter \(D\). The Colebrook-White equation is an implicit relationship that accounts for both roughness and Reynolds number \citep{colebrook1939}:
\begin{equation}
    \frac{1}{\sqrt{f}} = -2\log_{10}\left(\frac{\varepsilon}{3.7D} + \frac{2.51}{Re\sqrt{f}}\right).
\end{equation}

To simplify computations, explicit approximations to the Colebrook-White equation are commonly employed. We consider in this work the Swamee-Jain equation \citep{swamee1976}:
\begin{equation}
    f = 0.25 \left[\log_{10}\left(\frac{\varepsilon}{3.7D} + \frac{5.74}{Re^{0.9}}\right)\right]^{-2}.
\end{equation}

These equations are widely applied in engineering for estimating the friction factor in both smooth and rough pipes under turbulent flow conditions.

To characterize these properties, equations of state (EOS) are fundamental. They provide a mathematical relationship between key thermodynamic variables such as pressure (\(P\)), temperature (\(T\)), and density (\(\rho\)), enabling a comprehensive understanding of fluid behavior under varying flow conditions, particularly in the fields of petroleum engineering and chemical processing. While equations of state (EOS) typically depend on both \( P \) and \( T \), this study simplifies the analysis by considering density as a function of pressure only, assuming isothermal flow. For incompressible fluids, the EOS is trivial, as the density remains constant regardless of variations in pressure or temperature. Conversely, for compressible gases, we adopt the simplest model, assuming the ideal gas law:
\begin{equation}
    \rho = \frac{PM}{RT},
    \label{eq:eos_ideal_gas}
\end{equation}
where \(M\) is the molar mass of the gas, \(R\) is the universal gas constant, and \(T\) is the temperature.

In the petroleum industry, more complex equations of state (EOS) are commonly used to accurately model hydrocarbon mixtures under high-pressure and high-temperature conditions. Prominent examples include the Peng-Robinson EOS \citep{peng1976} and the Soave-Redlich-Kwong EOS \citep{soave1972}, which provide enhanced accuracy for real gases and reservoir fluids. These models are essential for predicting phase behavior, fluid properties, and thermodynamic equilibrium in petroleum engineering applications \citep{whitson2000}.

\subsection{Boundary Conditions}

Boundary conditions are crucial for defining the flow behavior at both the upstream and downstream ends of the system. Upstream boundary conditions can include several scenarios. One common boundary condition is the closed boundary, where the velocity is set to zero (\(V(x=0, t) = 0\)), indicating no flow entering the system. Another possibility is a pressure condition, where the upstream pressure is provided.

\begin{figure}[h!]
    \centering
    \includegraphics[width=1.15\textwidth, trim=150 100 60 100, clip]{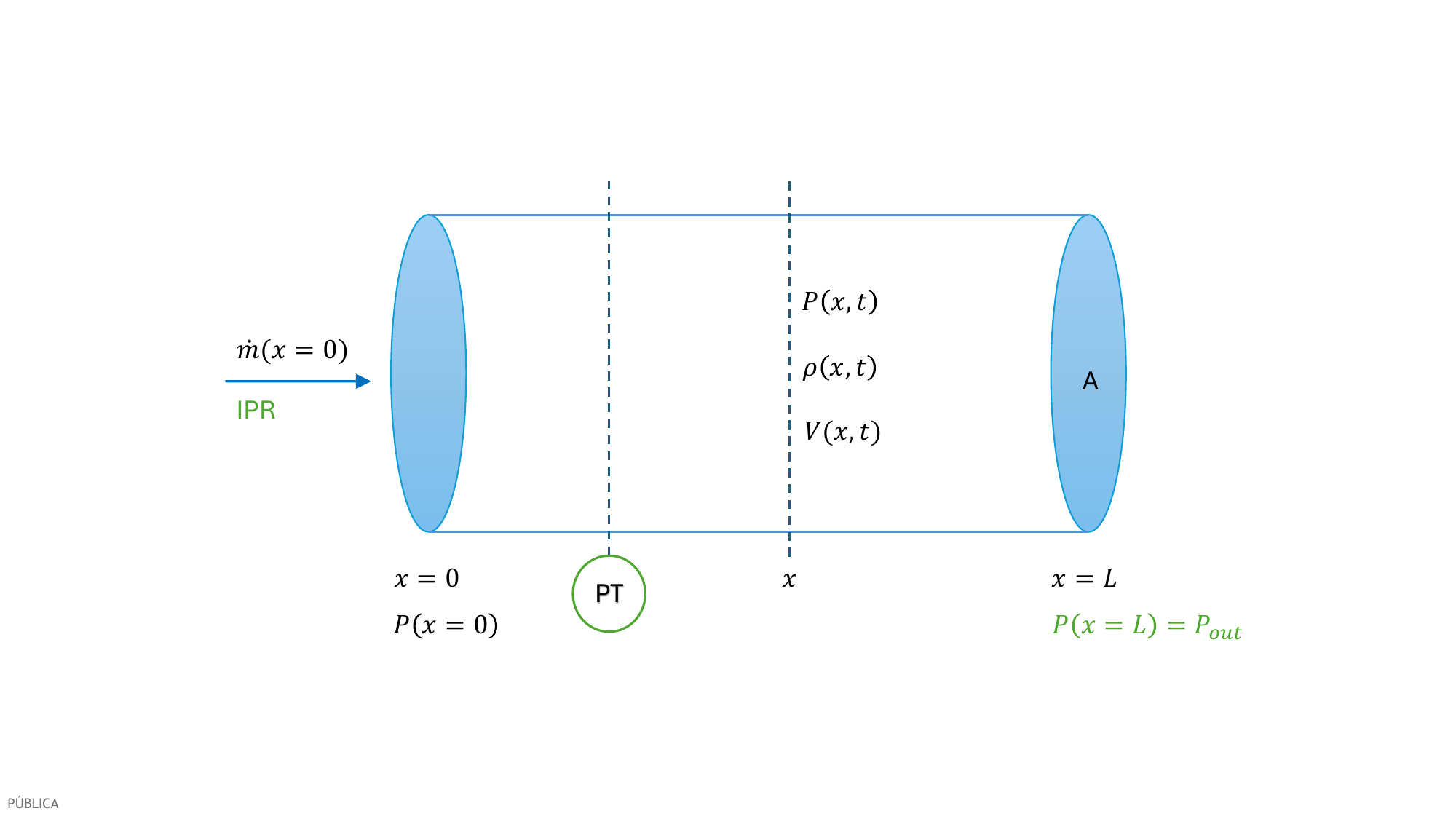}
    \caption{Schematic diagram illustrating the boundary conditions (highlighted in green), including the upstream IPR (Inflow Performance Relationship) and the downstream pressure condition. A pressure indicator and transmitter are shown at a specific point in the pipeline, representing the presence of a measurement sensor, commonly referred to as a Pressure Downhole Gauge (PDG). At an arbitrary point in the system, the state variables—pressure, velocity, and density—are highlighted. Temperature, which is also an important variable, is assumed constant in this study.
}
    \label{fig:variables}
\end{figure}

A more dynamic and realistic upstream condition is the Inflow Performance Relationship (IPR) \citep{aziz1979petroleum, vogel1968inflow}, which relates the mass flow rate \(\dot{m}(0, t)\) to the pressure drawdown. The linear IPR assumes that the mass rate is directly proportional to the pressure drawdown:

\begin{equation}
    \dot{m}(x=0, t) =  \text{PI} \left(P_{\text{reservoir}} - P(x=0, t)\right),
    \label{eq:IPR_condition_mass}
\end{equation}
where \(\text{PI}\) is the proportional constant (known as the productivity index), \(P_{\text{reservoir}}\) is the average pressure in the reservoir, which acts as a mass source supplying fluid to the system, and \(P(x=0, t)\) is the pressure at the upstream boundary.

This boundary condition reflects how the mass flow rate changes with the pressure difference, known as the drawdown. The larger the pressure difference between the reservoir and the downstream boundary, the higher the mass flow rate, representing the typical behavior of fluid entering a system such as a well or a pipeline.

Another option for the inlet boundary condition is to directly specify the mass flow rate. However, this approach does not typically capture the behavior of fluid flow in pipelines, particularly when restrictions at the production choke, located downstream, influence the system dynamics. The IPR-type boundary condition represents the flow reduction effect caused by valve restrictions, which is precisely the behavior we aim to capture for production control and optimization purposes.

By expressing the IPR condition in terms of mass flow rate, the boundary condition can be consistently applied across different flow regimes, including both compressible and incompressible flows. Additionally, this formulation aligns more naturally with practical field measurements, where mass flow rate is often the controlled or estimated variable at the wellhead or pipeline inlet.

The downstream boundary condition is set as follows:
\begin{equation}
    P(x=L, t) = P_{\text{out}}(t),
    \label{eq:fixed_pressure_condition}
\end{equation}
where \(P_{\text{out}}(t)\) represents the pressure at the outlet. The outlet pressure mimics the effect of choke opening through time, which is considered to be the manipulated variable (or control signal) of the system. Throughout this text, we will denote  \(P_{\text{out}}(t)\) as the control variable, written as $u(t)$:

\begin{equation*}
    P(x=L, t) = P_{\text{out}}(t) = u(t).
    \label{eq:fixed_pressure_condition_u}
\end{equation*}

\subsection{Incompressible Flow Modeling}

The modeling of incompressible water flow in pipelines is based on the conservation laws of mass and momentum. For the one-dimensional  case, the governing equations are derived from the fundamental mass transport and force balance formulations:
\begin{equation}
    \frac{\partial \rho}{\partial t} + \frac{\partial (\rho V)}{\partial x} = 0
    \label{eq:mass_conservation}
\end{equation}

Equation \ref{eq:mass_conservation} represents the conservation of mass in an one-dimensional flow. In this equation, \(\rho\) denotes the fluid density, \(V\) is the fluid velocity, \(t\) represents time, and \(x\) is the spatial coordinate along the direction of flow.

In the case of incompressible flow, the density \(\rho\) remains constant, and the mass conservation equation \eqref{eq:mass_conservation} simplifies to:
\begin{equation}
    \frac{\partial V}{\partial x} = 0
\end{equation}

This indicates that the velocity \(V\) does not vary along the flow direction for incompressible flow.

For an incompressible flow, where \( \frac{\partial V}{\partial x} = 0 \) and \( \rho \) is constant, the term \( \frac{\partial (\rho V^2)}{\partial x} \) vanishes. This is because:
\begin{equation}
    \frac{\partial (\rho V^2)}{\partial x} = \rho \frac{\partial (V^2)}{\partial x} = \rho \cdot 2V \frac{\partial V}{\partial x} = 0
\end{equation}

Therefore, the momentum equation \eqref{eq:momentum} simplifies significantly for incompressible flow. The simplified momentum equation becomes:
\begin{equation}\label{eq:momentum_inc}
    \rho \frac{\partial V}{\partial t} + \frac{\partial P}{\partial x} = -\rho g \sin\theta - \frac{1}{2} \rho f \frac{|V|V}{D}
\end{equation}

For the IPR upstream boundary condition, since the mass flow rate is proportional to the velocity and the density is constant, an IPR relationship can be expressed in terms of velocity instead of mass flow rate: 
\begin{equation}
    \dot{m} = \rho A V,
    \label{eq:mass_rate_definition}
\end{equation}
where \(A\) is the cross-sectional area and \(\rho\) is the fluid density, which remains constant for an incompressible fluid. This relationship shows that the mass flow rate is directly proportional to the velocity when the density and cross-sectional area are constant.

Therefore the velocity-based IPR boundary condition can be written as:
\begin{equation}
    V(x=0, t) = k \left(P_{\text{reservoir}} - P(x=0, t)\right),
    \label{eq:IPR_condition}
\end{equation}
where \(k\) is a proportional constant, \(P_{\text{reservoir}}\) is the reservoir pressure, and \(P(x=0, t)\) is the pressure at the upstream boundary.

In summary, the PDE system employed as loss functions for incompressible single-phase flow can be expressed as follows:
\begin{align*}
   &\mathcal{F}[V(x,t)] = \frac{\partial V}{\partial x} = 0, \quad x \in [0, L], \ t \in [0, T] 
\end{align*}
\begin{multline*}
   \mathcal{F}[V(x,t), P(x,t)] = \rho \frac{\partial V}{\partial t} + \frac{\partial P}{\partial x} + \rho g \sin \theta + \frac{1}{2} \rho f \frac{|V|V}{D} = 0, \\
      x \in [0, L], \ t \in [0, T]
\end{multline*}
\begin{align*}
   &\mathcal{B}[V(x,t), P(x,t)] = 
   \begin{cases} 
   V(0, t) - k (P_{\text{reservoir}} - P(0, t)) = 0, & x = 0, \ t \in [0, T] \\ 
   P(L, t) - P_{\text{out}}(t) = 0, & x = L, \ t \in [0, T]
   \end{cases}
\end{align*}
\begin{align*}
   &\mathcal{I}[V(x, 0), P(x, 0)] = 
   \begin{cases} 
   V(x, 0) - V_0(x) = 0, & \quad x \in [0, L], \\
   P(x, 0) - P_0(x) = 0, & \quad x \in [0, L].
   \end{cases}
\end{align*}
where \( V_0(x) \) \text{and} \( P_0(x) \) denote the known initial conditions ($t = 0$) for velocity and pressure, respectively, prescribed over the spatial domain \( x \in [0, L] \). Here, \(\mathcal{F}\), \(\mathcal{B}\), and \(\mathcal{I}\) represent, respectively, the dynamic equations, the boundary conditions, and the initial conditions.

\subsubsection{Normalized Equations for Incompressible Flow} \label{subsubsection:normalized_inc}

To ensure a balance between the mass and momentum equations during neural network training, it is important to achieve equilibrium in terms of their magnitudes. This balance is crucial for the training process to converge successfully. To facilitate this, both the inputs and outputs of the neural network are normalized, aiding in the efficiency and stability of the training process. The normalized variables are:

\[
\widetilde{t} = \frac{t}{t_\text{ref}}, \quad
\widetilde{x} = \frac{x}{x_\text{ref}}, \quad
\widetilde{V} = \frac{V}{V_\text{ref}}, \quad
\widetilde{P} = \frac{P}{P_\text{ref}}, \quad
\]

The governing equations for the normalized variables are presented below:
\begin{equation*}
\mathcal{F}[\widetilde{V}(\widetilde{x}, \widetilde{t})] = \frac{\partial \widetilde{V}}{\partial \widetilde{x}} = 0, \quad \widetilde{x} \in [0, 1], \, \widetilde{t} \in [0, 1]
\end{equation*}
\begin{multline}\label{eq:momentum_inc:B}
\mathcal{F}[\widetilde{V}(\widetilde{x}, \widetilde{t}), \widetilde{P}(\widetilde{x}, \widetilde{t})] = \frac{\partial \widetilde{V}}{\partial \widetilde{t}} + \frac{t_\text{ref} P_\text{ref}}{\rho V_\text{ref} x_\text{ref}} \frac{\partial \widetilde{P}}{\partial \widetilde{x}} + \frac{t_\text{ref} g \sin \theta}{V_\text{ref}} + \frac{1}{2} f \frac{t_\text{ref} V_\text{ref}}{D} |\widetilde{V}| \widetilde{V} = 0, \\ \widetilde{x} \in [0, 1], \, \widetilde{t} \in [0, 1]
\end{multline}
\begin{equation*}
\mathcal{B}[\widetilde{V}(\widetilde{x}, \widetilde{t}), \widetilde{P}(\widetilde{x}, \widetilde{t})] =
\begin{cases}
\widetilde{V}(0, \widetilde{t}) - k \left( \frac{P_\text{reservoir} - P_\text{ref} \widetilde{P}(0, \widetilde{t})}{V_\text{ref}} \right) = 0, & \widetilde{x} = 0, \, \widetilde{t} \in [0, 1] \\
\widetilde{P}(1, \widetilde{t}) - \frac{P_\text{out}}{P_\text{ref}} = 0, & \widetilde{x} = 1, \, \widetilde{t} \in [0, 1]
\end{cases}
\end{equation*}
\begin{equation*}
\mathcal{I}[\widetilde{V}(\widetilde{x}, 0), \widetilde{P}(\widetilde{x}, 0)] =
\begin{cases}
\widetilde{V}(\widetilde{x}, 0) - \frac{V_0(\widetilde{x})}{V_\text{ref}} = 0, & \widetilde{x} \in [0, 1] \\
\widetilde{P}(\widetilde{x}, 0) - \frac{P_0(\widetilde{x})}{P_\text{ref}} = 0, & \widetilde{x} \in [0, 1]
\end{cases}
\end{equation*}

For the steady-state regime, the term $\frac{\partial \widetilde{V}}{\partial \widetilde{t}}$ is zero in Equation \eqref{eq:momentum_inc:B} and the resulting equation for the momentum is:
\begin{align*}
\mathcal{F}[\widetilde{V}(\widetilde{x}, \widetilde{t}), \widetilde{P}(\widetilde{x}, \widetilde{t})] = \frac{ P_\text{ref}}{\rho V_\text{ref} x_\text{ref}} \frac{\partial \widetilde{P}}{\partial \widetilde{x}} + \frac{ g \sin \theta}{V_\text{ref}} + \frac{1}{2} f \frac{ V_\text{ref}}{D} |\widetilde{V}| \widetilde{V} = 0
\end{align*}

\subsection{Compressible Gas Flow}\label{subsection:compressible_gas_flow}

For compressible fluids, such as gas, which exhibit high compressibility, the modeling process becomes more complex, and the simplifications applied to incompressible systems are no longer valid. Therefore, the full mass and momentum conservation equations must be imposed, as defined in  Equations \eqref{eq:mass} and \eqref{eq:momentum}. Since density is a state variable, an equation of state (EOS) must be considered and, in our work, the ideal gas law is adopted, as shown in Equation \eqref{eq:eos_ideal_gas}. 

The boundary conditions are defined upstream by the mass-based IPR, according to Equations \eqref{eq:IPR_condition_mass} and \eqref{eq:mass_rate_definition}. For the downstream condition, the relationship established in Equation \eqref{eq:fixed_pressure_condition} is considered.

It is important to highlight that the equations for friction loss calculation are not explicitly formulated as loss functions. Instead, they are evaluated during the forward pass, based on the neural network's outputs. For example, the Reynolds Equation (\ref{eq:reynolds}) and the Blasius Equation (\ref{eq:blasius}) are computed sequentially within the forward pass, using the network’s predictions as input. Similarly, the equation of state (EOS) for the ideal gas law (\ref{eq:eos_ideal_gas}) is also applied in the forward pass, indicating that the density, while a state variable, is not directly predicted by the neural network.

Therefore, the outputs of the neural network in our formulation are only two: pressure and velocity. The density is computed as a function of pressure using the EOS (\ref{eq:eos_ideal_gas}), and the mass flow rate is obtained from the product of density and velocity (Equation \ref{eq:mass_rate_definition}). In the end, we have three state variables (pressure, density, and velocity) for two partial differential equations (mass and momentum conservation) and one algebraic equation (EOS), resulting in a fully determined system. Below, we represent the losses for the PDEs in terms of \( V(x,t) \) and \( P(x,t) \), but they are interchangeable. We could define the neural network output as \( \rho(x,t) \) and \( \rho(x,t) V(x,t)\), maintaining two outputs in the neural network, and the remaining state variables would be computed using the algebraic equation.

To sum up, the PDE system employed as loss functions for compressible single-phase flow can be stated as follows:
\begin{align*}
   &\mathcal{F}[V(x,t), P(x,t)] = \frac{\partial \rho}{\partial t} + \frac{\partial (\rho V)}{\partial x} = 0, \quad x \in [0, L], \ t \in [0, T]
\end{align*}
\begin{multline*}
   \mathcal{F}[V(x,t), P(x,t)] = \frac{\partial (\rho V)}{\partial t} + \frac{\partial (\rho V^2 + P)}{\partial x} +\rho g \sin\theta + \frac{1}{2} \rho f \frac{|V|V}{D} = 0, \\
   x \in [0, L], \ t \in [0, T]
\end{multline*}
\begin{multline*}
   \mathcal{B}[V(x,t), P(x,t)] = \\
   \begin{cases} 
   \rho(0, t) A V(0, t) - \text{PI}(P_{\text{reservoir}} - P(0, t)) = 0, & x = 0, \ t \in [0, T], \\ 
   P(L, t) - P_{\text{out}}(t) = 0, & x = L, \ t \in [0, T]
   \end{cases}
\end{multline*}
\begin{align*}
   &\mathcal{I}[V(x, 0), P(x, 0)] = 
   \begin{cases} 
   V(x, 0) - V_0(x) = 0, & \quad x \in [0, L], \\
   P(x, 0) - P_0(x) = 0, & \quad x \in [0, L].
   \end{cases}
\end{align*}
where \( V_0(x) \) \text{and} \( P_0(x) \) denote the known initial conditions ($t = 0$) for velocity and pressure, respectively, prescribed over the spatial domain \( x \in [0, L] \). Here, \(\mathcal{F}\), \(\mathcal{B}\), and \(\mathcal{I}\) represent, respectively, the dynamic equations, the boundary conditions, and the initial conditions.

\subsubsection{Normalized Equations for 
Compressible Flow} \label{subsubsection:normalized_comp}

As we did for the incompressible flow, we state the normalized variables as
\[
\widetilde{t} = \frac{t}{t_\text{ref}}, \quad
\widetilde{x} = \frac{x}{x_\text{ref}}, \quad
\widetilde{V} = \frac{V}{V_\text{ref}}, \quad
\widetilde{P} = \frac{P}{P_\text{ref}}, \quad
\widetilde{\rho} = \frac{\rho}{\rho_\text{ref}}.
\]

The loss functions with the normalized variables can be stated as follows:

\begin{align}\label{eq:mass_comp}
   &\mathcal{F}[\widetilde{V}(\widetilde{x},\widetilde{t}), \widetilde{P}(\widetilde{x},\widetilde{t})] = \frac{\partial \widetilde{\rho}}{\partial \widetilde{t}} + \frac{V_{\text{ref}} t_{\text{ref}}}{x_{\text{ref}}} \frac{\partial (\widetilde{\rho} \widetilde{V})}{\partial \widetilde{x}} = 0, \quad \widetilde{x} \in [0, 1], \ \widetilde{t} \in [0, 1]
\end{align}

\begin{multline}\label{eq:momentum_comp}
   \mathcal{F}[\widetilde{V}(\widetilde{x},\widetilde{t}), \widetilde{P}(\widetilde{x},\widetilde{t})] = \frac{\partial (\widetilde{\rho}\widetilde{V})}{\partial \widetilde{t}} 
   + \frac{V_{\text{ref}} t_{\text{ref}}}{x_{\text{ref}}} \frac{\partial (\widetilde{\rho}\widetilde{V}^2)}{\partial \widetilde{x}} 
   + \frac{P_{\text{ref}} t_{\text{ref}}}{\rho_{\text{ref}} V_{\text{ref}} x_{\text{ref}}} \frac{\partial \widetilde{P}}{\partial \widetilde{x}} \\
   + \frac{g t_{\text{ref}} \sin \theta}{V_{\text{ref}}} \widetilde{\rho} 
   + \frac{1}{2} f \left(\frac{V_{\text{ref}} t_{\text{ref}}}{D}\right) \widetilde{\rho} |\widetilde{V}| \widetilde{V} = 0, 
   \quad \widetilde{x} \in [0, 1], \ \widetilde{t} \in [0, 1]
\end{multline}
\begin{multline*}
   \mathcal{B}[\widetilde{V}(\widetilde{x},\widetilde{t}), \widetilde{P}(\widetilde{x},\widetilde{t})] =  \\
   \begin{cases} 
   \widetilde{\rho}(0, \widetilde{t}) \widetilde{V}(0, \widetilde{t}) - \frac{\text{PI}}{\rho_{\text{ref}} V_{\text{ref}} A}\left(P_{\text{reservoir}} - \widetilde{P}(0, \widetilde{t}) P_{\text{ref}}\right) = 0, & \widetilde{x}=0, \ \widetilde{t} \in [0, 1], \\[10pt] 
   \widetilde{P}(1, \widetilde{t}) - \frac{P_{\text{out}}}{P_{\text{ref}}} = 0, 
  & \widetilde{x}=1, \ \widetilde{t} \in [0, 1].
   \end{cases}
\end{multline*}
\begin{equation*}
\mathcal{I}[\widetilde{V}(\widetilde{x}, 0), \widetilde{P}(\widetilde{x}, 0)] =
\begin{cases}
\widetilde{V}(\widetilde{x}, 0) - \frac{V_0(\widetilde{x})}{V_\text{ref}} = 0, & \widetilde{x} \in [0, 1] \\
\widetilde{P}(\widetilde{x}, 0) - \frac{P_0(\widetilde{x})}{P_\text{ref}} = 0, & \widetilde{x} \in [0, 1]
\end{cases}
\end{equation*}

For the steady-state regime, Equations \eqref{eq:mass_comp} and \eqref{eq:momentum_comp} set the time derivative to zero, resulting in the  equations below:
\begin{align*}
   &\mathcal{F}[\widetilde{V}(\widetilde{x},\widetilde{t}), \widetilde{P}(\widetilde{x},\widetilde{t})] = \frac{\partial (\widetilde{\rho} \widetilde{V})}{\partial \widetilde{x}} = 0, \quad \widetilde{x} \in [0, 1]
\end{align*}
\begin{multline*}\label{eq:momentum_comp}
   \mathcal{F}[\widetilde{V}(\widetilde{x},\widetilde{t}), \widetilde{P}(\widetilde{x},\widetilde{t})] = \frac{V_{\text{ref}} }{x_{\text{ref}}} \frac{\partial (\widetilde{\rho}\widetilde{V}^2)}{\partial \widetilde{x}} 
   + \frac{P_{\text{ref}}}{\rho_{\text{ref}} V_{\text{ref}} x_{\text{ref}}} \frac{\partial \widetilde{P}}{\partial \widetilde{x}} \\
   + \frac{g \sin \theta}{V_{\text{ref}}} \widetilde{\rho} 
   + \frac{1}{2} f \left(\frac{V_{\text{ref}} }{D}\right) \widetilde{\rho} |\widetilde{V}| \widetilde{V} = 0, 
   \quad \widetilde{x} \in [0, 1]
\end{multline*}

\section{Related Works}

The numerical solution of single-phase flow problems involves addressing the coupling between pressure and velocity fields, which is critical to ensuring that the incompressibility condition is satisfied. 



In the SIMPLE (Semi-Implicit Method for Pressure-Linked Equations) algorithm introduced by Patankar and Spalding (1972) \citep{Patankar1972}, an iterative process is employed to address this coupling. The method starts with guessed values for the velocity and pressure fields. Using these initial guesses, the momentum equations are solved to obtain an updated velocity field. However, since this velocity field is based on an estimated pressure, it may not exactly satisfy the continuity equation.

To correct this pressure mismatch, the SIMPLE algorithm derives a pressure correction equation from the continuity equation. This correction equation is solved to obtain a pressure correction field, which is then used to adjust both the pressure and velocity fields iteratively, leading to a consistent and accurate solution for the coupled pressure-velocity system \citep{Malalasekera1995}. The SIMPLE method has evolved over time, with various extensions such as SIMPLEC (SIMPLE-Consistent) \citep{VanDoormal1984} and SIMPLER (SIMPLE-Revised) \citep{Patankar1979}, aiming to enhance convergence and stability. 


In our work, a semi-implicit finite difference scheme is implemented in a one-dimensional setting, where momentum conservation is handled using a staggered grid (half-cell offset) for the velocity components, following the approach introduced by \cite{Harlow1965}. The staggered grid arrangement offers the advantage of generating velocity values precisely at the cell faces required for scalar transport computations (convection-diffusion), eliminating the need for interpolation at the scalar cell faces \citep{Malalasekera1995}. This improves the accuracy and efficiency of the numerical scheme and reduces the occurrence of oscillating pressure fields, often known as the ``checkerboard'' effect.

The boundary conditions are specified by extrapolating the pressure field downstream, while an Inflow Performance Relationship (IPR) \citep{vogel1968inflow, aziz1979petroleum} is applied upstream. As we address one-dimensional flow systems with a modest number of grid points, a numerical strategy like SIMPLE was deemed unnecessary. Instead, the coupled nonlinear system between pressure and velocity was solved numerically to ensure conservation of the governing equations. The entire set of equations was directly solved using a nonlinear solver \citep{2020SciPy-NMeth}.


Following the discussion of traditional methods for solving the coupled mass and momentum systems using semi-implicit techniques, it is important to highlight recent advances in flow modeling achieved using physics-informed neural networks. These methods integrate physical laws directly into the training process of neural networks, allowing for the efficient and accurate solution of flow-related problems while leveraging both data and governing equations to maintain physical consistency. 

Several recent studies have explored fluid flow modeling using Physics-Informed Neural Networks (PINNs). These works have demonstrated the application of PINNs to solve problems governed by partial differential equations, such as the Navier-Stokes equations, in both compressible and incompressible flow regimes \citep{RAISSI2019686, mao2020highspeed, jin2021nsfnets,  kharazmi2021hpvpinns}. Some studies employ physics-informed neural networks (PINNs) in more applied contexts, such as gas pipelines \citep{ZHANG2023106073} and multiphase oil and gas flow, specifically in the dynamic modeling of electrical submersible pump (ESP) systems \citep{Carvalho_2024}.

Building upon the success of PINNs in modeling systems governed by PDEs, recent research \citep{mowlavi2022optimalcontrolpdesusing, barrystraume2022physicsinformedneuralnetworkspdeconstrained, Faria2024} has expanded their application to address control problems, including both classical optimal control and reinforcement learning approaches. While traditional PINNs have primarily focused on solving PDEs that describe system dynamics, such as the Navier-Stokes equations, incorporating control objectives within the PINN framework has opened new avenues for optimizing system performance. By embedding control variables and objectives directly into the loss function alongside governing physical laws, these methods enable simultaneous learning of system dynamics and optimal control strategies. 

\cite{mowlavi2022optimalcontrolpdesusing} and \cite{barrystraume2022physicsinformedneuralnetworkspdeconstrained} solve the traditional PINN losses for PDEs simultaneously, considering initial conditions (IC), boundary conditions (BC), and the governing PDEs, along with the losses concerning the cost functional, which accounts for the control law and the resulting solution. These approaches integrate the system state solution and the optimal control trajectory into a single learning framework. 
  However, for practical applications such as real-time monitoring and control, it would be necessary to retrain the network from scratch for any new target, making this approach less feasible for real-time scenarios where quick adjustments are needed.

The main advantage of our approach, particularly for real-time control and monitoring systems, builds on PINC \citep{ANTONELO2024127419}, where the control input and initial states are treated as inputs to the neural network. This allows for long-term simulation, enabling optimal control strategies to be determined using Model Predictive Control (MPC) techniques over successive time windows. Once the networks are trained, the neural network can be used to compute the optimal trajectory to reach a target without requiring retraining. 

The key difference between our 
approach and the PINC framework in \cite{ANTONELO2024127419} is that we are now addressing more complex systems governed by partial differential equations (PDEs) rather than ordinary differential equations (ODEs). This expansion significantly broadens the model's scope, making it applicable to fluid flow problems, where both spatial and temporal dynamics must be considered.


%% file: 4-methods.tex
\section{Methods}

In this section, we explore Physics-Informed Neural Networks (PINNs) as surrogates for modeling flows governed by partial differential equations (PDEs), specifically the mass and momentum conservation laws, although our proposal can be applied to any system described by PDE as long as certain assumptions hold, which are presented later in this section.

We develop two different PINNs: one for the steady-state regime, where the partial derivatives with respect to time are neglected, and another for the transient regime, where the full conservation equations are taken into account. 
The steady-state PINN models the system behavior at equilibrium, while the transient PINN captures the time-dependent dynamics of the flow. 
\eric{Note that the former is instrumental for training the latter, as it will be shown in Section~\ref{sec:pinc_t}}.

\subsection{Physics-Informed Neural Networks (PINNs)}
\label{sec:pinn}

In this paper, we consider nonlinear PDEs of the following general form:
\begin{equation}
   \partial_{\tilde{t}} \mathbf{y}(\tilde{x},\tilde{t}) + \mathcal{N}[\mathbf{y}(\tilde{x},\tilde{t})] = 0, \quad \tilde{t} \in [0,1], \quad \tilde{x} \in [0,1], \label{eq:general_pde}
\end{equation}
where $\mathcal{N}[\cdot]$ is a nonlinear differential operator, and $\mathbf{y}(\tilde{x},\tilde{t})$ represents the state of the dynamic system, depending on both normalized time \(\tilde{t}\) and normalized spatial coordinate \(\tilde{x}\).

This system is subject to boundary conditions:
\begin{equation}
   \mathcal{B}[\mathbf{y}(\tilde{x},\tilde{t})] = 0, \quad \tilde{x} \in \{0, 1\}, \quad \tilde{t} \in [0, 1], \label{eq:boundary_conditions}
\end{equation}
and initial conditions:
\begin{equation}
   \mathcal{I}[\mathbf{y}(\tilde{x},0)] = 0, \quad \tilde{x} \in [0, 1]. \label{eq:initial_conditions}
\end{equation}

We define $\mathcal{F}(\mathbf{y})$ as representing the left-hand side of Equation (\ref{eq:general_pde}):
\begin{equation}
\mathcal{F}(\mathbf{y}) := \partial_{\tilde{t}} \mathbf{y}(\tilde{x},\tilde{t}) + \mathcal{N}[\mathbf{y}(\tilde{x},\tilde{t})] 
\label{eq:pde}
\end{equation}

In this context, $\mathbf{y}(\tilde{x},\tilde{t})$ denotes the output of a deep neural network, where $\mathbf{y} = f_\mathbf{w}(\tilde{x},\tilde{t})$ represents the mapping learned by the neural network. The function $f_\mathbf{w}(\tilde{x},\tilde{t})$ is parameterized by a set of weights $\mathbf{w}$ that are adjusted during training.

The neural network is trained using optimizers such as ADAM \citep{Kingma2014} or L-BFGS \citep{Andrew2007} to minimize a mean squared error (MSE) cost function:

\begin{equation}
\textrm{MSE} = \lambda_{\mathcal{D}} \cdot \textrm{MSE}_{\mathcal{D}} + \lambda_{\mathcal{F}} \cdot \textrm{MSE}_{\mathcal{F}} + \lambda_{\mathcal{B}} \cdot \textrm{MSE}_{\mathcal{B}} + \lambda_{\mathcal{I}} \cdot \textrm{MSE}_{\mathcal{I}},
\label{eq:erro_pde}
\end{equation}

\allowdisplaybreaks
\begin{subequations}
\begin{align}
\textrm{MSE}_{\mathcal{D}} &= \frac{1}{N_y} \sum_{i=1}^{N_y}  \frac{1}{N_{\mathcal{D}}} \sum_{j=1}^{N_{\tilde{x},\tilde{t}}} | y_i(\tilde{x}^j,\tilde{t}^j) - \widehat{y}_i^j |^2,  \label{eq:data_loss_pde} \\
\textrm{MSE}_{\mathcal{F}} &=  \frac{1}{N_{ge}} \sum_{i=1}^{N_{ge}} 
   \frac{1}{N_{\mathcal{F}}} \sum_{k=1}^{N_{\mathcal{F}}}  | \mathcal{F}_i( \mathbf{y}(\tilde{x}^k,\tilde{t}^k) ) |^2, \label{eq:pde_loss_pde} \\
\textrm{MSE}_{\mathcal{B}} &=  \frac{1}{N_{bc}} \sum_{i=1}^{N_{bc}} 
   \frac{1}{N_{\mathcal{B}}} \sum_{l=1}^{N_{\mathcal{B}}}  | \mathcal{B}_i( \mathbf{y}(\tilde{x}^l,\tilde{t}^l) ) |^2,
\label{eq:bc_loss_pde} \\
\textrm{MSE}_{\mathcal{I}} &=  \frac{1}{N_{ic}} \sum_{i=1}^{N_{ic}} 
   \frac{1}{N_{\mathcal{I}}} \sum_{m=1}^{N_{\mathcal{I}}}  | \mathcal{I}_i( \mathbf{y}(\tilde{x}^m,0) ) |^2, \label{eq:ic_loss_pde}
\end{align}
\end{subequations}
in which  $N_{\mathcal{D}}$, $N_{\mathcal{F}}$, $N_{\mathcal{B}}$, $N_{\mathcal{I}}$, and $N_y$ correspond to the number of training data points (measured data), collocation points for the PDE, boundary condition points, initial condition points, and the number of neural network outputs, respectively; $y_i(\tilde{x},\tilde{t})$ is the $i$-th output of the neural network, $\widehat{y}_i^j$ represents the corresponding observed data value at point $(\tilde{x}^j, \tilde{t}^j)$, \textcolor{black}{$\mathbf{y}(\tilde{x}^l,\tilde{t}^l)$ is the boundary value at $(\tilde{x}^l,\tilde{t}^l)$, and $\mathbf{y}(\tilde{x}^m,0)$ is the initial condition at $(\tilde{x}^m,0)$.}

Moreover, $N_{ge}$, $N_{bc}$, and $N_{ic}$ denote the counts of governing equations (e.g., 2 for mass and momentum conservation), number of boundary conditions imposed and initial conditions, respectively. For the problem stated in Section \ref{subsection:general_modeling}, $N_y = N_{ge} = N_{bc} = N_{ic} = 2$.

The terms in the loss function are weighted by the parameters \( \lambda_{\mathcal{D}} \), \( \lambda_{\mathcal{F}} \), \( \lambda_{\mathcal{B}} \), and \( \lambda_{\mathcal{I}} \) to adjust the relative importance of each component. The parameter \( \lambda_{\mathcal{D}} \) controls the importance of fitting the model to the observed data points. The parameter \( \lambda_{\mathcal{F}} \) regulates how well the model adheres to the governing partial differential equations (PDEs), ensuring that the solution respects the physical laws. The parameter \( \lambda_{\mathcal{B}} \) balances the satisfaction of the boundary conditions, while \( \lambda_{\mathcal{I}} \) ensures that the solution satisfies the initial conditions.

\subsection{PINNs for Control (PINC)} \label{sec:pinc_intro}

\eric{Physics-Informed Neural Nets for Control (PINC), introduced in \cite{ANTONELO2024127419}, extends the traditional PINN framework by incorporating control variables and the range of initial conditions, enabling their use in control applications. However, the original PINC formulation is limited to learning ODE solutions. In that work, two additional input dimensions—the control signal \( \tilde{u} \) and the initial state—are introduced alongside the continuous time input, allowing a PINN to be used for control in continuous time for the first time. 
To handle variable long-range simulations, which traditional PINNs cannot achieve, PINC operates autoregressively over shorter time intervals. In this process, each prediction serves as feedback: the initial state (input) for the next interval is set to the final predicted state (output) of the previous interval, improving accuracy in extended simulations.}

\eric{For PDE systems, PINC must be extended to include an additional input, the spatial dimension \( \tilde{x} \). However, this leads to a significant increase in input dimensionality, as the initial state now becomes high-dimensional. To address this, the PINC for PDEs proposed in this work replaces the full initial state input with the control signal from the previous time window, reducing input dimensionality considerably. Assuming the system reaches a steady state under a constant control signal within each time window, this simplification enhances training efficiency while maintaining accuracy.
}

\begin{figure}[H]
    \centering
    \includegraphics[width=1.0\textwidth,trim=2cm 6.0cm 3.5cm 6cm,clip]{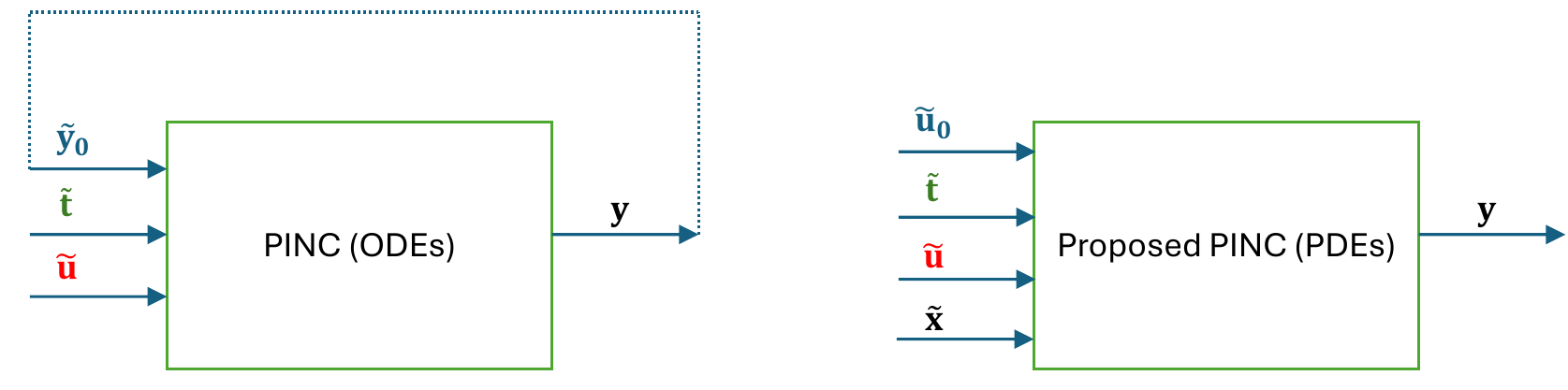}
    \caption{Proposed schematic illustrating the extension of the original PINC, initially designed for ODEs, to the approach presented in this work---a PINN architecture tailored for control applications. The proposed architecture differs from the original PINC in two key aspects: the inclusion of the spatial coordinate (as we are now dealing with PDEs) and the treatment of the initial condition. Unlike the original approach, the initial condition does not directly incorporate the initial states \( \tilde{y}_0 \), as this would significantly increase the input dimensionality of the network due to the initial condition being a function of the spatial coordinate \( \tilde{x} \). To address this, we propose a simplified parameterization of the initial condition based on the control input \( \tilde{u}_0 \), which drives the plant to steady-state conditions. Thus, the system is assumed to start from the steady-state regime achieved under \( \tilde{u}_0 \), and the Proposed PINC for PDEs captures the system dynamics over space and time as the current control input \( \tilde{u} \) is applied.
}
    \label{fig:pinc_diffs}
\end{figure}

The PINC presented in Figure~\ref{fig:pinc_diffs} accounts for transient behavior and will be referred to as PINC-Transient throughout this text. A simplified version of this neural network for the steady-state regime, referred to as PINC Steady-State, will also be introduced, as explained in the next section (Section~\ref{sec:pinc_ss}). Naturally, for the steady-state regime, the time component \( \tilde{t} \) and the initial conditions \( \tilde{u}_0 \) are no longer included as inputs to the neural network, as the system is assumed to reach a stable equilibrium regardless of the initial conditions.

\subsection{PINNs for Control (PINC) in the Steady-State Regime}\label{sec:pinc_ss}

In the steady-state regime, where temporal derivatives are neglected, PINC models the flow system considering a wide range of downstream pressure values, represented as control inputs. 
\eric{Starting with the PINN model presented in Section~\ref{sec:pinn} with the time $t$ input removed},
a second input dimension, \( \tilde{u} \), is introduced alongside the existing spatial dimension \( \tilde{x} \), which represents the normalized position in the system. While \( \tilde{x} \) captures the spatial variation of the system variables (pressure, velocity, and density), \( \tilde{u} \) represents the normalized control value influencing the downstream boundary condition (downstream pressure). This allows the PINC to generalize over multiple control settings, adapting to a wide range of operational conditions.

In the steady-state regime, the governing equations are:
\begin{equation}
   \mathcal{N}[\mathbf{y}(\tilde{x},\tilde{u})] = 0, \quad \tilde{x} \in [0,1], \quad \tilde{u} \in \mathbb{R}, \label{eq:general_pde_steady}
\end{equation}

This system is subject to boundary conditions:
\begin{equation}
   \mathcal{B}[\mathbf{y}(\tilde{x},\tilde{u})] = 0, \quad \tilde{x} \in \{0, 1\}. \label{eq:boundary_conditions_steady}
\end{equation}

We define $\mathcal{F}(\mathbf{y})$ as representing the left-hand side of Equation (\ref{eq:general_pde_steady}):
\begin{equation}
\mathcal{F}(\mathbf{y}) := \mathcal{N}[\mathbf{y}(\tilde{x},\tilde{u})] 
\label{eq:pde_steady}
\end{equation}

In this context, $\mathbf{y}(\tilde{x},\tilde{u})$ denotes the output of a deep neural network.
More explicitly, $\mathbf{y} = f_\mathbf{w}(\tilde{x},\tilde{u})$ represents the neural network mapping from $\tilde{x},\tilde{u}$ to the PDE solution, $\mathbf{y}$,
parametrized by $\mathbf{w}$.  
The neural network is trained using optimizers such as ADAM \citep{Kingma2014} or L-BFGS \citep{Andrew2007} to minimize 
the following MSE cost function:
\begin{equation}
\textrm{MSE} = \lambda_{\mathcal{F}} \cdot \textrm{MSE}_{\mathcal{F}} + \lambda_{\mathcal{B}} \cdot \textrm{MSE}_{\mathcal{B}},
\label{eq:erro_pde_steady}
\end{equation}
    
\begin{align}
\textrm{MSE}_{\mathcal{F}} &= \frac{1}{N_{ge}} \sum_{i=1}^{N_{ge}} 
   \frac{1}{N_{\mathcal{F}}} \sum_{k=1}^{N_{\mathcal{F}}}  | \mathcal{F}_i( \mathbf{y}(\tilde{x}^k,\tilde{u}^k) ) |^2, \label{eq:pde_loss_pde_steady}
\end{align}

%

\begin{align}
\textrm{MSE}_{\mathcal{B}} &= \frac{1}{N_{bc}} \sum_{i=1}^{N_{bc}} 
   \frac{1}{N_{\mathcal{B}}} \sum_{l=1}^{N_{\mathcal{B}}}  | \mathcal{B}_i( \mathbf{y}(\tilde{x}^l,\tilde{u}^l) ) |^2 \nonumber, \label{eq:bc_loss_pde_steady_aligned}
\end{align}



There is no initial condition loss because we are focusing on the steady-state regime, where the system reaches an equilibrium solution that is independent of the initial conditions.

The points \( (\tilde{x}, \tilde{u}) \) used for training are generated using the Latin Hypercube Sampling (LHS), a statistical method designed to efficiently sample multidimensional parameter spaces \citep{McKay1979}. 

For the PDE loss, we generate \( N_{\mathcal{F}} \) collocation points \( (\tilde{x}^k, \tilde{u}^k ) \), where each pair is sampled from the interval \([0,1]\) using a bidimensional LHS approach. For the boundary condition loss, we generate \( N_{\mathcal{B}} \) points using a unidimensional LHS approach for the control input \( \tilde{u}^{l} \). The position $\tilde{x}^l$ is either 0 or 1, depending on how the boundary condition is defined for the $i$-th boundary condition equation.


\subsection{PINNs for Control (PINC) in the Transient Regime}\label{sec:pinc_t}

\eric{In this section,}
the PINC model is designed to handle dynamical systems governed by partial differential equations (PDEs), where the state evolution depends on both spatial and temporal variables, as well as control inputs. The goal of the transient PINC model is to learn the underlying dynamics of the system and serve as a surrogate model for optimal control strategies.

The transient PINC model operates by dividing the total simulation time into smaller time windows, during which the control input is held constant, as shown in Figure \ref{fig:free_simulation}. Each time window represents a fixed interval of time over which the system dynamics are predicted by the neural network. Within a given time window, the model assumes that the control signal \(\tilde{u}\) does not change, allowing the network to focus on predicting the system's state for the duration of the window. After each time window is completed, the control input can be updated, and the process is repeated for the next window. This approach ensures that the control strategy can be adjusted dynamically over time.

The inputs to the transient PINC model are the normalized spatial coordinate \(\tilde{x}\), the normalized time variable \(\tilde{t}\), the control input at the previous time window \(\tilde{u}_0\), and the control input at the current time window \(\tilde{u}\).

The state of the system, $\mathbf{y}(\tilde{x}, \tilde{t}, \tilde{u}_0, \tilde{u})$, is predicted by the neural network, which maps the inputs to the system's output. The goal is to minimize the total loss function, which consists of three main components: the PDE loss, the boundary condition loss (BC loss), and the initial condition loss (IC loss).

The system is governed by a nonlinear PDE of the following form:
\begin{equation}
\partial_{\tilde{t}} \mathbf{y}(\tilde{x}, \tilde{t}, \tilde{u}_0, \tilde{u}) + \mathcal{N}[\mathbf{y}(\tilde{x}, \tilde{t}, \tilde{u}_0, \tilde{u})] = 0, \quad \tilde{x} \in [0,1], \quad \tilde{t} \in [0,1],
\end{equation}
where $\mathcal{N}[\cdot]$ is a nonlinear differential operator that describes the dynamics of the system. The PDE loss is defined as the mean squared error of the PDE residuals, evaluated at a set of collocation points $(\tilde{x}^k, \tilde{t}^k, \tilde{u}_0^k, \tilde{u}^k)$. This is mathematically expressed as:
\begin{equation}
\textrm{MSE}_{\mathcal{F}} = \frac{1}{N_{ge}} \sum_{i=1}^{N_{ge}} 
   \frac{1}{N_{\mathcal{F}}} \sum_{k=1}^{N_{\mathcal{F}}}  \left | \mathcal{F}_i( \mathbf{y}(\tilde{x}^k,\tilde{t}^k, \tilde{u}_0^k, \tilde{u}^k) ) \right |^2,
\label{eq:pde_loss_pde_transient}
\end{equation}

The boundary conditions are applied at the spatial boundaries of the system, $\tilde{x} = 0$ and $\tilde{x} = 1$, representing the upstream and downstream conditions, respectively. It is formulated as:

\begin{align}
\textrm{MSE}_{\mathcal{B}} = \frac{1}{N_{bc}} \sum_{i=1}^{N_{bc}} \left( \frac{1}{N_{\mathcal{B}}} \sum_{l=1}^{N_{\mathcal{B}}} \left | \mathcal{B}_i( \mathbf{y}(\tilde{x}^l, \tilde{t}^l, \tilde{u}^l_0, \tilde{u}^l) ) \right|^2 \right),
\label{eq:bc_loss_pde_transient}
\end{align}

Since the system evolves over time, we impose an initial condition at $\tilde{t} = 0$. The initial condition $y(\tilde{x}, 0, \tilde{u}_0, \tilde{u})$ must match the steady-state solution obtained from the previously trained steady-state PINC model. The assumption rests on the fact that the time window is large enough to reach stability with the control $\tilde{u}_0$, which simplifies the input space of the neural network. This simplification will be further elaborated in the following Section (\ref{ssec:simplifying_ic}). The IC loss is defined as:
\begin{equation}
\textrm{MSE}_{\mathcal{I}} = \frac{1}{N_y} \sum_{i=1}^{N_y} \frac{1}{N_{\mathcal{I}}} \sum_{m=1}^{N_{\mathcal{I}}} \left | \mathbf{y}_i(\tilde{x}^m, 0, \tilde{u}_0^m, \tilde{u}^m) - \bar{\mathbf{y}}_i^{SS}(\tilde{x}^m, \tilde{u}_0^m) \right |^2, 
\label{eq:ic_loss_pinc_transient}
\end{equation}
where $\bar{y}_i^{SS}(\tilde{x}^m, \tilde{u}_0^m)$ represents the steady-state solution obtained from the previously trained (with fixed weights) steady-state PINC model, as described in Section \ref{sec:pinc_ss}. 
The parameter $N_{\mathcal{I}}$ denotes the number of points used to impose the initial condition. 
\eric{Eq. (\ref{eq:ic_loss_pinc_transient}) represents the connection between Steady-State (SS) PINC and the transient PINC (Fig.~\ref{fig:flowchart_IC}), where the former is used to train the latter by generating the IC targets}.

The total loss function for the transient PINC model is a weighted sum of the PDE loss, the BC loss, and the IC loss, being expressed as:
\begin{equation}
\textrm{MSE} = \lambda_{\mathcal{F}} \cdot \textrm{MSE}_{\mathcal{F}} + \lambda_{\mathcal{B}} \cdot \textrm{MSE}_{\mathcal{B}} + \lambda_{\mathcal{I}} \cdot \textrm{MSE}_{\mathcal{I}},
\label{eq:total_loss_pinc_transient}
\end{equation}
where $\lambda_{\mathcal{F}}$, $\lambda_{\mathcal{B}}$, and $\lambda_{\mathcal{I}}$ are weighting factors that control the relative importance of each loss component.

\subsubsection{Sampling Strategy and Training Process}

The collocation points for the PDE loss in Equation \eqref{eq:pde_loss_pde_transient} are generated using Latin hypercube sampling \citep{McKay1979} in a four-dimensional space, with each point being a tuple $(\tilde{x}, \tilde{t}, \tilde{u}_0, \tilde{u})$. This approach ensures that the sample points are well-distributed across the input space, such that:
\[
(\tilde{x}, \tilde{t}, \tilde{u}_0, \tilde{u}) \in [0,1]^4.
\]

The points for the BC loss in Equation \eqref{eq:bc_loss_pde_transient} are generated independently for each boundary using LHS in three dimensions, with each point being a tuple $(\tilde{t}, \tilde{u}_0, \tilde{u})$ sampled within the interval:
\[
(\tilde{t}, \tilde{u}_0, \tilde{u}) \in [0,1]^3.
\]

The points for the IC loss in Equation \eqref{eq:ic_loss_pinc_transient} are generated using LHS in three dimensions, with each point being a tuple $(\tilde{x}, \tilde{u}_0, \tilde{u})$ sampled within the interval:
\[
(\tilde{x}, \tilde{u}_0, \tilde{u}) \in [0,1]^3.
\]
This ensures that the initial condition is enforced for different spatial positions and control inputs, matching the steady-state solution from the previous time window as initial condition across a variety of configurations.

The transient PINC model is trained considering the total MSE loss (Equation \ref{eq:total_loss_pinc_transient}) using optimization algorithms such as ADAM \citep{Kingma2014} and L-BFGS \citep{Andrew2007}.

\subsubsection{Simplifying Assumption for the Initial Condition}\label{ssec:simplifying_ic} 

In developing the PINC model for the transient regime, a simplifying assumption was made regarding the initial condition to facilitate training and reduce the model's complexity. In previous PINC models designed for ODE-based systems \citep{ANTONELO2024127419,Kittelsen:ASOC:2024}, the dynamics were modeled as \(y = f(\tilde{t}, \tilde{x}_0, \tilde{u})\), where \(\tilde{x}_0\) represents the initial condition of the dynamic system. Here, we adopt a different approach to the initial condition, using the control input in the previous time window \(\tilde{u}_0\) as an input feature to the neural network to handle the system's initialization.

The transient PINC model is designed to capture the temporal evolution of the system's dynamics within a time window of length \(T\), conditioned on an initial state that is equal to the steady-state solution obtained under the control \(\tilde{u}_0\) applied in the previous time window. This assumption is valid if the time window length \(T\) is sufficiently large for the system to reach stability. 

Including \(\tilde{u}_0\) as an input to the neural network eliminates the need to model a wide range of initial conditions that would otherwise depend on the spatial variable \(\tilde{x}\), thereby reducing the dimensionality of the input space. Consequently, the neural network's predictive capability is enhanced and the training process becomes more efficient. 

If the model were to use the actual states \(\tilde{x}_0\) directly as input, generating multiple initial conditions that vary spatially would be needed to train the network, complicating the model's structure and substantially increasing its input dimensionality. 
This added complexity would likely require the use of dimensionality reduction techniques to maintain a manageable model size, such as the application of Koopman embeddings \citep{Geneva_2022}.

Also, random sampling strategies such as LHS would not provide the training inputs for the NN with a realistic family of initial conditions, since the states are spatially correlated according to the governing equations (\ref{subsection:general_modeling}), unlike the PINC originally proposed in \cite{ANTONELO2024127419} for ODEs.

Although this is a simplification, it is a highly appropriate one, as many transient simulation applications in practice consider the steady-state regime as the initial condition. This assumption simplifies the model's ability to capture the system's dynamics by focusing on a family of realistic initial conditions. 
Many flow simulations in real-world scenarios begin from an established steady state, such as in production shutdowns in oil and gas wells, pipeline startup sequences, or reservoir pressure buildups following a long shut-in period \citep{10.2118/109162-MS}. Additionally, processes like hydraulic fracturing and gas-lift operations also often rely on steady-state conditions as the initial point for simulating transient behavior.

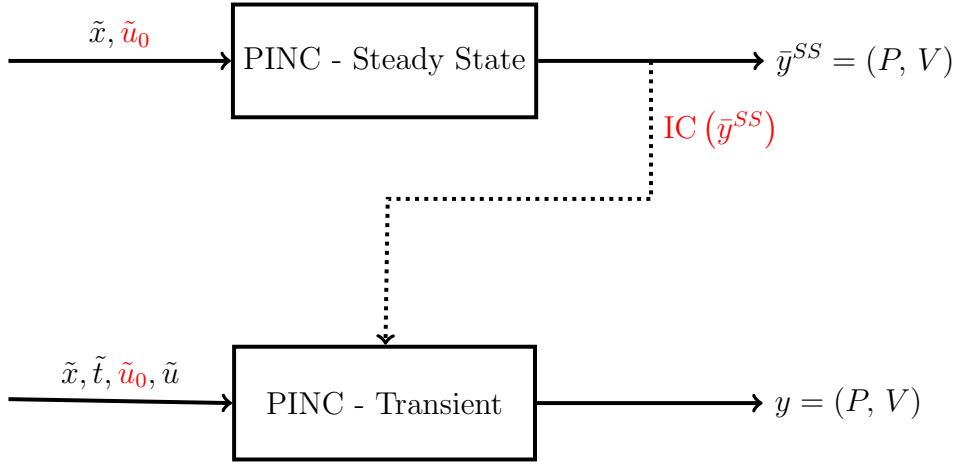
\begin{figure}[h!]
\centering
\begin{tikzpicture}[node distance=3cm, auto, thick]

    \node (steady) [draw, rectangle, line width=1.5pt, minimum width=4cm, minimum height=1.5cm, text centered] {PINC - Steady State};
    \node (transient) [draw, rectangle, line width=1.5pt, below=3cm of steady, minimum width=4cm, minimum height=1.5cm, text centered] {PINC - Transient};

    \draw[->, line width=1.5pt] (-5, 0) -- (steady.west) node[midway, above] {$\tilde{x}, \textcolor{red}{\tilde{u}_0}$};
    \draw[->, line width=1.5pt] (-5, -4.5) -- (transient.west) node[midway] {$\tilde{x}, \tilde{t}, \textcolor{red}{\tilde{u}_0}, \tilde{u}$};

    \draw[->, line width=1.5pt] (steady.east) -- node[pos=0.5, above](icstart){} ++(3,0) node[right] {$\bar{y}^{SS} = (P,\,V)$};

    \draw[->, line width=1.5pt] (transient.east) -- ++(3,0) node[right] {$y = (P,\,V)$};

    \draw[->, line width=1.5pt, dotted]
        (icstart) 
        -- ++(0,-2.0) node[midway, right, text=red] {$\text{IC}\left(\bar{y}^{SS}\right)$}
        -- ++(-3.5,0) 
        -- (transient);

\end{tikzpicture}
\caption{\eric{Flowchart illustrating the use of predicted initial conditions (IC) from the steady-state PINC solution in the training of the transient PINC.}
Note that the Steady-State PINC is trained in a first stage, and subsequently, the Transient PINC is trained using the predictions of the former that represent the equilibrium solution $\bar{y}^{SS}$ for the respective inputs $\tilde{x}$ and $\tilde{u}_0$.}
\label{fig:flowchart_IC}
\end{figure}

As illustrated in the flowchart (Figure \ref{fig:flowchart_IC}), during the training process 
\eric{of the transient PINC,} the \eric{already trained} steady-state PINC model (Section \ref{sec:pinc_ss}) provides the equilibrium solution for the state variables, \(\bar{y}^{SS}\), based on the control \(\tilde{u}_0\). These equilibrium solutions are then used as the initial conditions (IC) for the transient PINC model, \eric{as shown in the Equation (\ref{eq:ic_loss_pinc_transient}) for the IC Loss}. The transient model takes the normalized inputs \(\tilde{x}\), \(\tilde{t}\), \(\tilde{u}_0\), and \(\tilde{u}\) to simulate the system's dynamic evolution over time, starting from the steady-state condition achieved in the previous time window. 
\eric{Therefore, we replace the initial conditions \(\tilde{x}_0\) used in \cite{ANTONELO2024127419} with \(\tilde{u}_0\), which serves as an input feature representing the initial condition in the PINC framework of this work, as a simplifying assumption.}

\subsubsection{Forward Simulation}\label{subsubsection:forward_simulation}

Until now, we have not explicitly specified how the variables are indexed to their respective time windows. The control \( \tilde{u} \) represents the signal applied in the current time window, whereas \( \tilde{u}_0 \) corresponds to the control signal from the previous window.

\begin{figure}[htbp]
    \centering
    \includegraphics[width=1.0\textwidth]{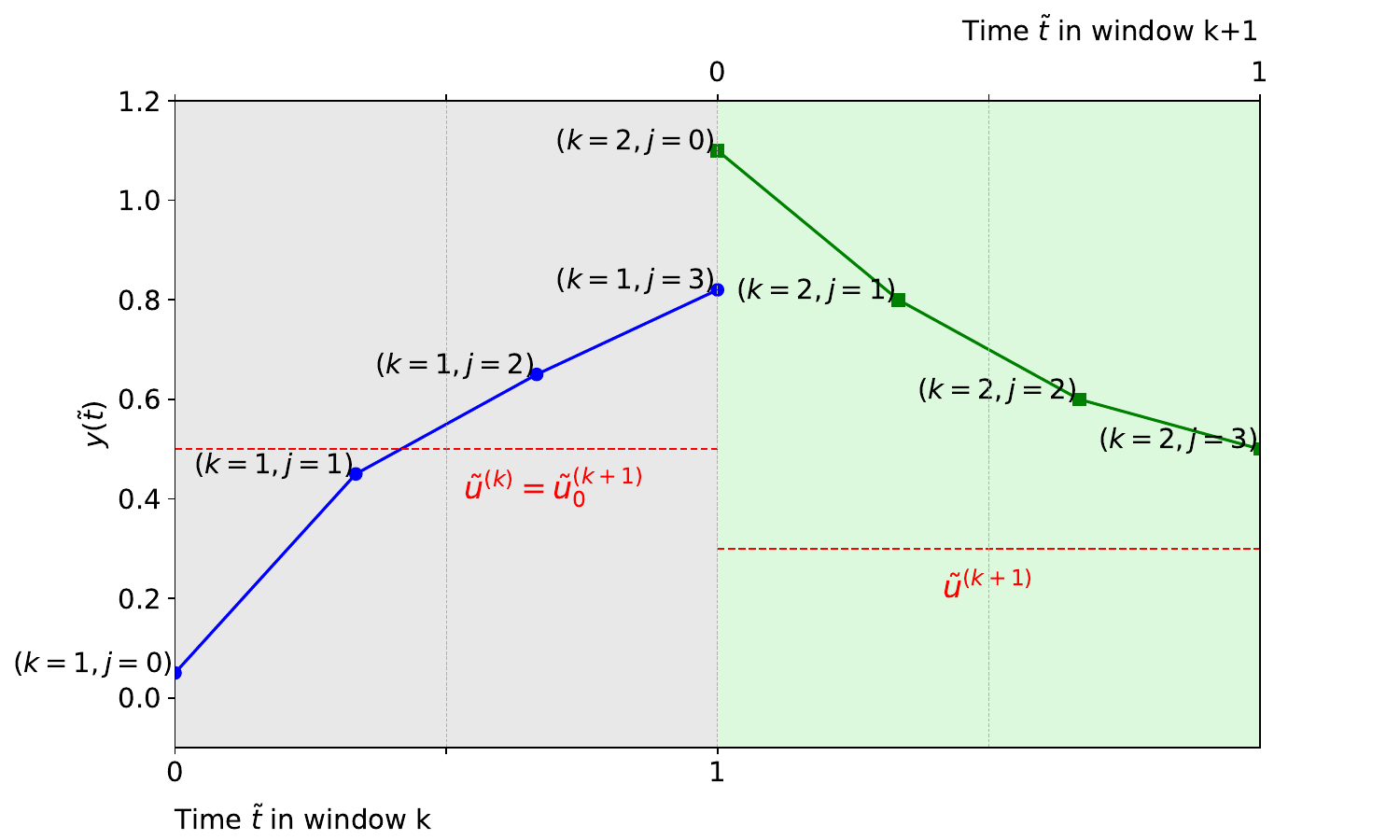}
    \caption{Transient PINC schematic representation of \eric{two consecutive} temporal windows for a given position $\tilde{x}$, \eric{with $M=3$, \textit{i.e.}, 4 timesteps inside each window}.
    The \eric{predictions}     \( y^{(k, j)} \) for the time window indexed by \( k \) are shown in blue, while \( y^{(k+1, j)} \) for the subsequent time window \( k+1 \) are depicted in green.
    \eric{Both blue and green points are predictions of the PINC network, while the edges between them were just plotted. These edges could be smoothed if more PINC predictions are computed for the intermediate time steps.}
    This scheme is a visual representation of the simulation in Algorithm \ref{alg:forward-sim}, where the control signal $\tilde{u}^{(k)}$ is held constant for each time window $k$, \eric{shown as horizontal red dashed lines}. Note that the PINC accepts continuous inputs, and can predict for any intermediate 
    $\tilde{t}$ value.
    \eric{Furthermore, 
    \( y^{(k=1, j=3)} \) is not necessarily equal to \( y^{(k=2, j=0)} \), as shown by the mismatch of the last blue point and the first green point.
    }
    }

    \label{fig:free_simulation}
\end{figure}

To clarify the evolution of the dynamical system across different time windows, we introduce a more specific notation, aligned with Figure \ref{fig:free_simulation}, which illustrates the computation of the forward simulation over these time windows. The forward simulation progresses using two indices: \( k \) for the time window and \( j \) for the discrete time step within each window.

The index \( k \) represents the temporal window, where \( k = 1, 2, \dots, N \), with each window corresponding to a specific segment of the overall simulation period during which the control signal remains constant. 
Window \( k \) covers the time interval \([(k-1)T, kT]\), where \( T \) denotes the duration of each window.

The index \( j \) denotes the discrete time step within each window, where \( j = 0, 1, \ldots, M-1 \), and discretizes the time within each window \( k \). The neural network's output at window \( k \) and time step \( j \), denoted by \( y^{(k, j)} \), represents the state variable at a given spatial location $\tilde{x}$.

The system's dynamics are described by the neural network function 
\( f(\tilde{x}, \tilde{t}_j, \tilde{u}_0^{(k)}, \tilde{u}^{(k)}) \), where \( \tilde{x} \) represents the spatial position, 
\( \tilde{t}_j = \frac{j}{M-1} \times \frac{T}{t_{\text{ref}}} \) denotes the normalized time within window \( k \), ranging from \( 0 \) to \( \frac{T}{t_{\text{ref}}} \), 
\( \tilde{u}_0^{(k)} \) corresponds to the control from the previous window \( k-1 \), and \( \tilde{u}^{(k)} \) represents the control applied in the current window \( k \). 
The output of the neural network at time step \( j \) within window \( k \), denoted as \( y^{(k,j)} \), is given by the following equation:

\begin{equation}
    y^{(k,j)}(\tilde{x}) = f(\tilde{x}, \tilde{t}_j, \tilde{u}_0^{(k)}, \tilde{u}^{(k)}).
\end{equation} 
\eric{Note that $\tilde{t}_j$ can be any value in  $[0,1]$, representing continuous time, even though the notation used here discretizes this time with the $j$ index in $M+1$ time steps within time window $k$, which is useful for performance evaluation and plotting purposes.}
To simplify the notation, we will denote \( y^{(k,j)}(\tilde{x}) \) simply as \( y^{(k,j)} \). The Forward Simulation can be more easily understood through the code of Algorithm \ref{alg:forward-sim}.

It is important to note in Algorithm \ref{alg:forward-sim} that there is no auto-regressive feedback in the model, meaning that the output at the beginning of a new time window, \( y^{(k+1, 0)} \), is not the same as the final output of the previous window, \( y^{(k, M-1)} \). \textit{This implies that any errors made during one time window do not accumulate or propagate to the next as time progresses.} Instead, the initial condition for time window \( k \) relies solely on the control value \( u_0^{(k)} \), as explained in Section \ref{ssec:simplifying_ic}. In other words, the state in each window is initialized based on the control applied in the previous window, but there is no direct dependency between the outputs across windows.

\begin{algorithm}[H]
\caption{Forward Simulation of the System}\label{alg:forward-sim}
\KwData{Initial control $\tilde{u}_0^{(1)}$, sequence of controls $\{\tilde{u}^{(1)}, \tilde{u}^{(2)}, \ldots, \tilde{u}^{(N)}\}$, number of windows $N$, number of time steps per window $M$}
\KwResult{Outputs $y^{(k, j)}$ for all windows $k = 1, 2, \ldots, N$ and time steps $j = 0, 1, \ldots, M-1$}
\For{$k = 1$ \KwTo $N$}{
    \For{$j = 0$ \KwTo $M-1$}{
        Compute the output: 
        \[
        y^{(k, j)} = f(\tilde{x}, \tilde{t}_j, \tilde{u}_0^{(k)}, \tilde{u}^{(k)})
        \]
        where $\tilde{t}_j = \frac{j}{M-1} \times \frac{T}{t_{\text{ref}}}$.
    }
    Update the initial control for the next window: 
    \[
    \tilde{u}_0^{(k+1)} = \tilde{u}^{(k)}
    \]
}
\end{algorithm}

Figure \ref{fig:free_simulation} provides a schematic representation of the temporal windows in the forward simulation process. The neural network’s output values, \( y^{(k, j)} \), representing the state variables for window \( k \), are shown in blue, while the values \( y^{(k+1, j)} \) for window \( k+1 \) are displayed in green. It is worth noting that \( y^{(k=1, j=3)} \neq y^{(k=2, j=0)} \), as there is no auto-regressive feedback in the model. Furthermore, the initial condition for time window \( k+1 \), represented by the point \( y^{(k=2, j=0)} \), depends only on the control value \( \tilde{u}^{(k)} = \tilde{u}_0^{(k+1)} \).

\subsection{Model Predictive Control}

The predictive control approach using the PINC model involves deriving a sequence of control actions \( \tilde{u}_1, \tilde{u}_2, \ldots, \tilde{u}_N \) that guide the system toward a desired target state. Starting from an initial state conditioned by the control input \( \tilde{u}_0 \), the optimization process aims to find a smooth progression of control actions that will gradually steer the system toward the target. This is achieved by leveraging the fixed weights of the pre-trained PINC model for the transient regime.

\subsubsection{Model Predictive Control using PINC Transient} \label{subsubsection:mpc_Transient}

The objective function for the predictive control problem is formulated as follows:
\begin{equation}
\begin{aligned}
\min_{\tilde{u}_1, \tilde{u}_2, \dots, \tilde{u}_{N_c}} ~~& \sum_{i=1}^{N_p} 
      \left( f(\bar{x}, \tilde{T}_s, \tilde{u}_{i-1}, \tilde{u}_i) - y_{\text{target}}(\bar{x}) \right)^2
+ \lambda \sum_{i=1}^{N_c} \left( \tilde{u}_i - \tilde{u}_{i-1} \right)^2
   \\
\text{subj. to:} ~& f(\bar{x}, \tilde{T}_s, \tilde{u}_0, \tilde{u}_1) - y_0 \leq \Delta y_{\text{max}} \\
& y_0 - f(\bar{x}, \tilde{T}_s, \tilde{u}_0, \tilde{u}_1) \leq \Delta y_{\text{max}} \\
& f(\bar{x}, \tilde{T}_s, \tilde{u}_i, \tilde{u}_{i+1}) - f(\bar{x}, \tilde{T}_s, \tilde{u}_{i-1}, \tilde{u}_i) \leq \Delta y_{\text{max}}, ~ \forall i = 1,  \dots, N_p-1 \\
& f(\bar{x}, \tilde{T}_s, \tilde{u}_{i-1}, \tilde{u}_i) - f(\bar{x}, \tilde{T}_s, \tilde{u}_i, \tilde{u}_{i+1}) \leq \Delta y_{\text{max}}, ~ \forall i = 1,  \dots, N_p-1 \\
& \tilde{u}_i = \tilde{u}_{N_c}, \quad \forall i = N_c+1, \dots, N_p
\end{aligned}
\label{eq:mpc_transient}
\end{equation}
\textcolor{black}{where $\Delta y_{\text{max}}\geq |f(\bar{x}, \tilde{T}_s, \tilde{u}_{i}, \tilde{u}_{i+1}) - f(\bar{x}, \tilde{T}_s, \tilde{u}_{i-1}, \tilde{u}_{i}) |$ limits the maximum variation of the controlled output.}

In the PINC Transient case, \( f(\bar{x}, \tilde{t}, \tilde{u}_{i-1}, \tilde{u}_i) \) represents the transient output from the PINC model (Section \ref{sec:pinc_t}) for the controlled variable (\textit{e.g.}, pressure) at a fixed position \( \bar{x} \) and time \( \tilde{t} \). Its initial condition is denoted by \( y_0 \). The fixed position \( \bar{x} \) in our flow system corresponds to the location where measurements from the PDG sensor are available, as depicted in Figure \ref{fig:variables}.

The objective function minimizes the deviation of the controlled variable from the target value \( y_{\text{target}}(\bar{x}) \), while incorporating a penalty term \( \lambda \) on control variations between intervals. The selection of the target value for this problem will be further discussed in the results section (\ref{subsubsection:mpc_inc}).

The control horizon (\(N_c\)) represents the number of control inputs treated as independent decision variables. For \(i > N_c\), the control inputs are fixed at \(\tilde{u}_{N_c}\) to reduce computational complexity, meaning that the control horizon is smaller than the prediction horizon.

The prediction horizon (\(N_p\)) defines the number of future time steps over which the system's behavior is predicted, with longer horizons improving accuracy at the cost of increased computation. The normalized sampling time ($\tilde{T}_s$) specifies the interval between successive control updates.

The hard constraint  $\Delta y_{\text{max}}\geq |f(\bar{x}, \tilde{T}_s, \tilde{u}_{i}, \tilde{u}_{i+1}) - f(\bar{x}, \tilde{T}_s, \tilde{u}_{i-1}, \tilde{u}_{i}) |$ limits the maximum allowable change in the neural network's output between consecutive intervals. This variation is evaluated at each sampling time (\( \tilde{t} = \tilde{T_s} \)), providing the MPC controller with dynamic restrictions that govern the system's behavior over time.

The optimization problem described above (Equation \ref{eq:mpc_transient}) is solved using the CasADi framework \citep{andersson2019casadi}, which provides a symbolic environment for optimization and dynamic system modeling. In this context, the nonlinearities introduced by the PINC model are handled through automatic differentiation, allowing the solver to compute the necessary gradients for the optimization process. The solver employed is IPOPT (Interior Point OPTimizer) \citep{wachter2006ipopt}, a robust tool for solving large-scale Nonlinear Programming (NLP) problems. IPOPT is well-suited for handling the nonlinear constraints present in this MPC formulation.

The workflow in CasADi involves defining the optimization variables, the objective function, and the constraints symbolically. CasADi then automatically computes the Jacobians and Hessians required by IPOPT to solve the problem efficiently, providing an optimal sequence of control inputs.

\begin{figure}[h!]
    \centering
    \includegraphics[width=1.0\textwidth, trim=60 150 60 100, clip]{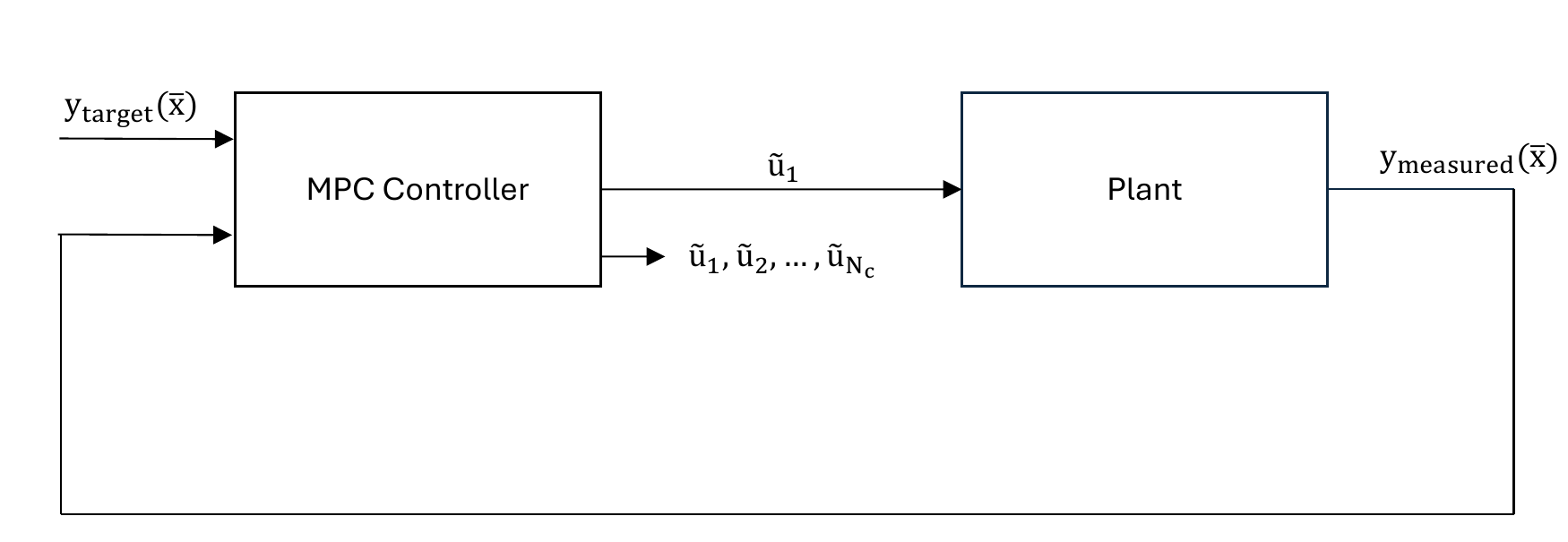}
    \caption{Visual scheme of the real-time MPC control acting on the plant (simulated through a numerical scheme) according to Algorithm \ref{alg:mpc_closed_loop}. The MPC controller relies on its predictive model based on the transient PINC and on feedback from the available measurements at $\bar{x}$, a position typically represented by the PDG (pressure downhole gauge).}

    \label{fig:mpc_scheme}
\end{figure}

It is worthwhile mentioning, as part of best practices in engineering, that the MPC Controller operates in a closed-loop configuration with the plant, as shown in Figure \ref{fig:mpc_scheme}. In this work, the plant represents the system solving the partial differential equations using a numerical finite difference scheme.

The reference position ($\bar{x}$) is assumed to be known and fixed, providing observable and measurable values from the plant. These measured values are used as feedback for the controller, enabling it to recalibrate its outputs based on the model prediction (PINC) and the collected data. This approach ensures dynamic interaction between the controller and the plant, as detailed in Algorithm \ref{alg:mpc_closed_loop}.

In step 5 of Algorithm \ref{alg:mpc_closed_loop}, the predictive model is refined by incorporating the error between the measured output, \( y_\text{measured} \), and the predictions from the PINC Transient model. This adjustment is based on a feedback mechanism to improve the model’s accuracy, as discussed in \cite{jordanou_filter}. \\

\begin{algorithm}[H]
\label{alg:mpc_closed_loop}
\caption{MPC Control Algorithm with Plant Feedback}\label{alg:mpc_feedback}
\KwData{Initial control $\tilde{u}_0$, prediction horizon $N_p$, control horizon $N_c$, sampling time $T_s$, maximum allowable change ($\Delta y_\text{{max}}$) number of iterations $N$}
\KwResult{Control actions $\tilde{u}_1, \tilde{u}_2, \ldots, \tilde{u}_N$ applied to the plant}

Initialize: $t \gets 0$\; 

\For{$k = 1$ \KwTo $N$}{
    Obtain the current output $y_{\text{measured}}(t)$\;

    Set $y_0 \gets y_{\text{measured}}(t)$\; 
    
    Solve the optimization problem described in (\ref{eq:mpc_transient}) obtaining the optimal control sequence $\tilde{u}_1, \tilde{u}_2, \ldots, \tilde{u}_{N_c}$\;
    
    Apply $\tilde{u}_1$ to the plant for duration $T_s$\;
    
    Update time: $t \gets t + T_s$\;

    Set $\tilde{u}_0 \gets \tilde{u}_1$\; 
}
\end{algorithm}

%% file: 5-results.tex
\section{Experiments and Analysis} \label{sec:results}

\subsection{Incompressible Flow}\label{subsection:inc_flow_results}

This section handles the flow problem under the assumptions of isothermal and incompressible flow. The parameters used for the simulation of the incompressible single-phase water flow system are summarized in Table~\ref{tab:simulation_parameters}. The pipe diameter (\(D\)) was set to 0.1 meters, and the fluid viscosity (\(\mu\)) was 0.001 Pa·s, representing typical values for water. The IPR parameter (\(k\)) was chosen as \(1 \times 10^{-5}\). The table also presents reference values for pressure and velocity, which are used in the normalization of balance and momentum governing equations, presented in Section \ref{subsubsection:normalized_inc}.

The static pressure (\(P_{\text{reservoir}}\)) at the upstream boundary was set to \(2 \times 10^5\) Pa, and the total length of the pipe was 100 meters. Additionally, the fluid density (\(\rho\)) was assumed to be 1000 kg/m\(^3\), consistent with the properties of water under incompressible flow conditions.

\begin{table}[h!]
    \centering
        \caption{Simulation parameters used for the incompressible and horizontal single-phase water flow system, including MPC controller settings (3 last rows).}
    \label{tab:simulation_parameters}
    \begin{tabular}{|l|c|c|}
        \hline
        \textbf{Parameter} & \textbf{Symbol} & \textbf{Value} \\ \hline
        Pipe Diameter & \(D\) & 0.1 m \\ \hline
        Fluid Viscosity & \(\mu\) & 0.001 Pa·s \\ \hline
        IPR Parameter & \(k\) & \(1 \times 10^{-5}\) \\ \hline
        Reservoir Pressure & \(P_{\text{reservoir}}\) & \(2 \times 10^5\) Pa \\ \hline
        Total Pipe Length ($x_{\text{ref}}$)& - & 100 m \\ \hline
        Inclination & \(\theta\) & 0 \\ \hline
        Pressure Reference & \(P_{\text{ref}}\) & \(1 \times 10^5\) Pa \\ \hline
        Velocity Reference & \(V_{\text{ref}}\) & 1 m/s \\ \hline
        Time Reference & \(t_{\text{ref}}\) & 10 s \\ \hline
        
        Fluid Density & \(\rho\) & 1000 kg/m\(^3\) \\ \hline
        Friction Factor Calculation & \(f\) & Blasius Equation\\ \hline        
        Control Horizon & \(N_c\) & 2 \\ \hline
        Prediction Horizon & \(N_p\) & 10 \\ \hline
        Sampling Time & \(T_s\) & 1 s \\ \hline
    \end{tabular}

\end{table}

\subsubsection{PINC in the Steady-State Regime}

\begin{figure}[h!]
    \centering
    \includegraphics[width=1.0\textwidth]{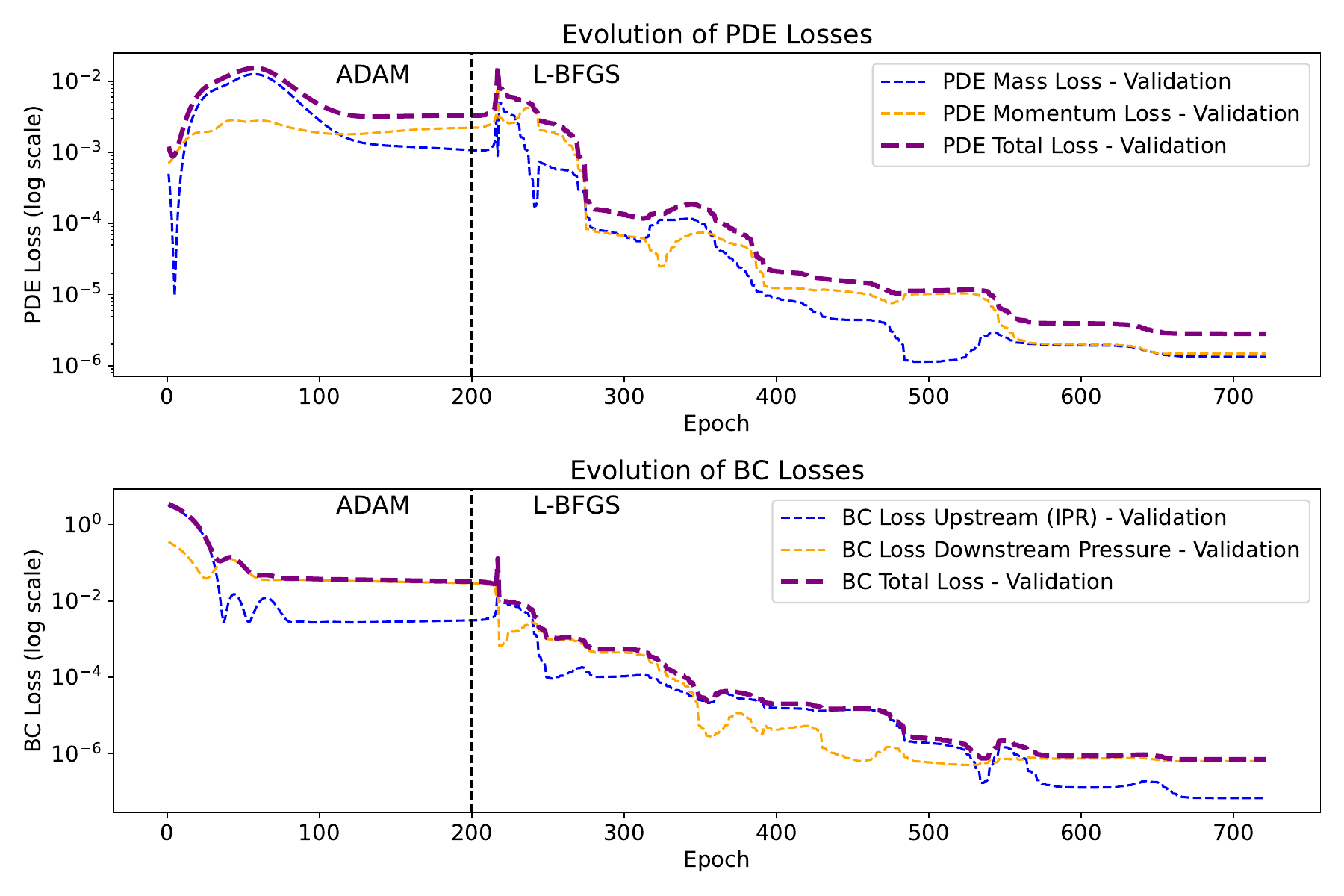}
    \caption{The training process of the PINC in the steady-state framework highlights the importance of normalizing the governing mass and momentum equations, ensuring a balanced contribution from both terms. This normalization allows the network to converge to a sufficiently small validation loss (on the order of $10^{-6}$) over the epochs. Additionally, a similar equilibrium is observed in the boundary condition losses, further attributed to the applied normalization. The rationale behind this normalization for both the governing equations and boundary conditions losses is detailed in Section \ref{subsubsection:normalized_inc}. The training begins with 200 epochs using the ADAM optimizer, followed by further refinement with the L-BFGS optimizer, which requires additional iterations to achieve improved performance.}
    
    \label{fig:SS_training}
\end{figure}

For the steady-state PINN model described in Section~\ref{sec:pinc_ss}, we employed a fully connected neural network (MLP) with 4 hidden layers, each containing 20 neurons. The hyperbolic tangent (\(\tanh\)) activation function was used for all layers. Optimization was performed in two stages: the first 200 iterations utilized the ADAM optimizer to initialize the parameters, followed by the L-BFGS optimizer for fine-tuning, as shown in Figure \ref{fig:SS_training}. 

The total number of collocation points used for the PDE loss is \(N_{\mathcal{F}} = 1000\). For the boundary condition loss, \(N_{\mathcal{B}} = 200\) points were utilized, evenly distributed between the two boundaries: the upstream boundary, where the IPR equation is imposed, and the downstream boundary, where the pressure is specified. Since the model predicts two outputs (pressure and velocity), the number of outputs is \(N_y = 2\).

A key aspect of the implementation is the computation of the Reynolds number, which plays a critical role in defining the friction factor in the momentum equation. To ensure numerical stability and prevent the propagation of extreme values, the Reynolds number is constrained using the \texttt{torch.clamp} function. Specifically, the Reynolds number is limited to a range such that a non-zero lower bound for the Reynolds number is ensured. This clamping mechanism is essential for stabilizing the training process, as it prevents overflow or underflow in the computation of the friction factor \(f\).

The normalization strategy presented in Section \ref{subsubsection:normalized_inc} was carefully chosen to ensure that the terms \(\frac{\partial \widetilde{V}}{\partial \widetilde{x}}\) and \(\frac{\partial \widetilde{V}}{\partial \widetilde{t}}\) remain compatible in scale. This is achieved because both the numerator (neural network outputs) and denominator (neural network inputs) of these terms are bounded, ensuring consistency in their magnitudes. This property is particularly important as these terms directly contribute to the residuals of the governing equations, maintaining balance between the components of the loss function. By adopting this normalization, we aim to achieve compatibility in terms of magnitude across the components of the loss function, preventing any single term from dominating due to scale differences.

The same principle applies to the boundary and initial condition losses, where \(\widetilde{V}\) and \(\widetilde{P}\) appear explicitly. Both quantities are normalized using reference values \(V_{\text{ref}}\) and \(P_{\text{ref}}\), which ensures that their contributions to the loss function are well-behaved and consistent in scale. By construction, this normalization leverages the fact that the chosen reference values inherently limit \(\widetilde{V}\) and \(\widetilde{P}\), facilitating stable and effective training of the model. Figure \ref{fig:SS_training} illustrates this property, demonstrating the balance achieved between the mass and momentum equations, as well as the consistency of the boundary conditions imposed at each boundary.

Having achieved a promising validation loss for the PINC model in the steady-state regime, the primary objective is to evaluate the neural network's capability to predict the spatial distribution of output variables, such as pressure and velocity, under steady-state equilibrium conditions. To achieve this, we analyze the behavior of the profiles for different values of the control variable \(\widetilde{u}\), which represents the normalized outlet pressure. 

Since the flow is assumed to be incompressible, the velocity profile is expected to remain constant along the spatial domain for each \(\widetilde{u}\), as dictated by the mass conservation principle \(\left(\frac{dV}{dx} = 0\right)\). The pressure profile is derived from Equation (\ref{eq:momentum_inc}) neglecting the partial derivative with respect to time. This implies the following equation for the steady-state incompressible flow regime:
\begin{equation}\label{eq:momentum_inc_2}
     \frac{\partial P}{\partial x} = -\rho g \sin\theta - \frac{1}{2} \rho f \frac{|V|V}{D}
\end{equation}

This is, in fact, a nonlinear differential equation in which the pressure loss is computed by accounting for both gravitational and frictional contributions. Since the velocity remains constant for a given \(\widetilde{u}\), the Reynolds number also remains constant (as the density, diameter, and viscosity are assumed to be fixed). Consequently, the friction factor is constant, resulting in a uniform pressure gradient \(\left(\frac{\partial P}{\partial x}\right)\) for a given control signal in the steady-state regime.

\begin{figure}[h!]
    \centering
    \includegraphics[width=1.0\textwidth]{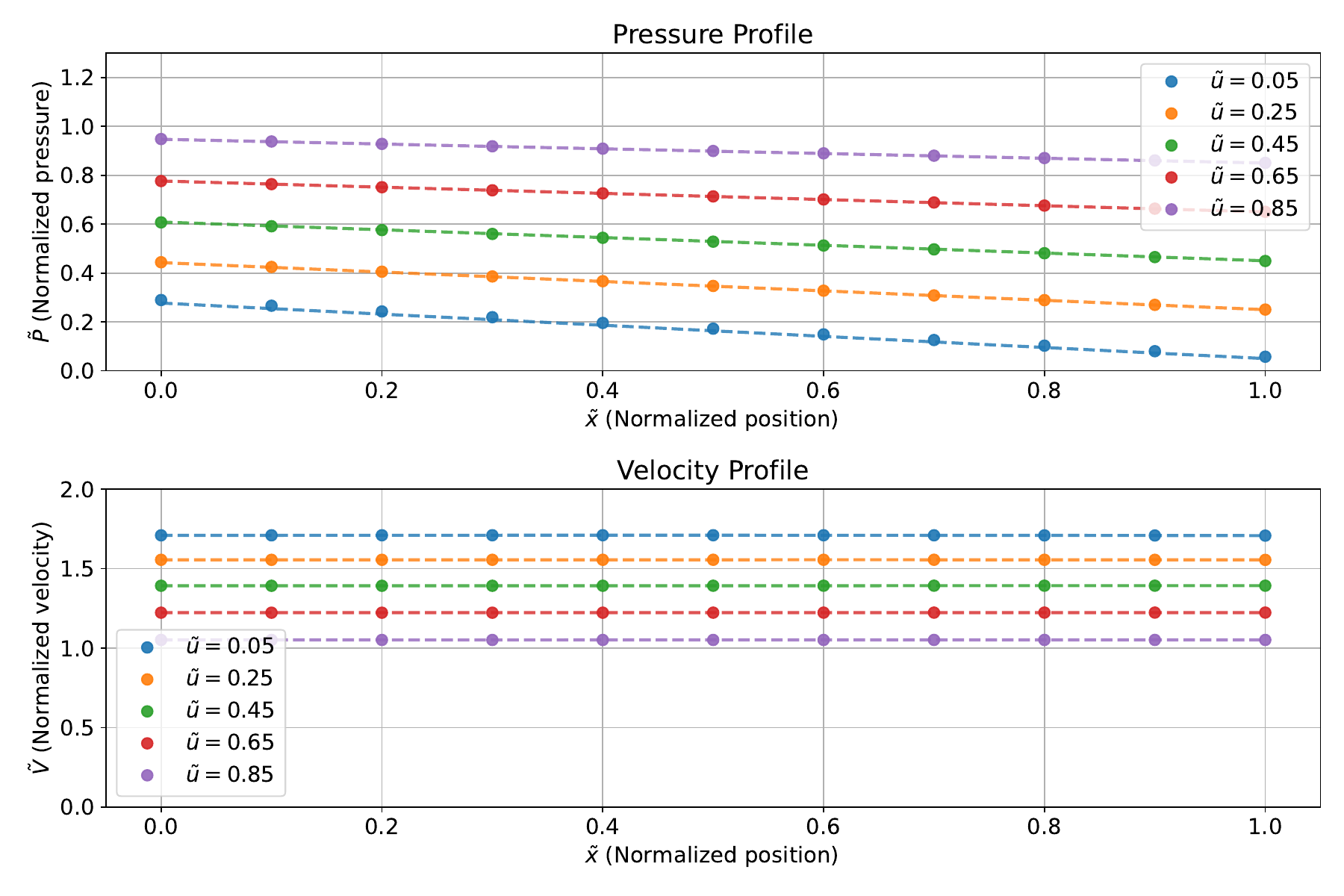}
    \caption{Comparison between numerically simulated results (dashed line) and those computed by the PINC (solid points). The Steady-State PINC performs as expected, capturing the spatial trend of pressure and velocity distributions across a range of control values ($\widetilde{u}$). These results support the use of the Steady-State PINC model as an initial condition estimator (as outlined in Section \ref{ssec:simplifying_ic}) for training the Transient PINC model.} 

    \label{fig:simulation_pinc_ss:incompressible}
\end{figure}

The pressure and velocity profiles in the steady-state regime are shown in Figure \ref{fig:simulation_pinc_ss:incompressible}, comparing the numerically simulated results with those predicted by the PINC model. As expected, both the velocity profiles and pressure gradients remain constant, highlighting the neural network's strong adherence to the underlying physics. This consistency aligns with the validation losses depicted in Figure \ref{fig:SS_training}, where the mass and momentum losses are on the order of magnitude of \(10^{-6}\).

This observation underscores one of the key strengths of the formulation presented in Section~\ref{sec:problem_statement}, namely the incorporation of the IPR as a boundary condition. By adopting this approach, the velocity is intrinsically determined as part of the problem formulation rather than being imposed as an explicit boundary condition. Specifically, given an external control input \(\widetilde{u}\), the neural network seamlessly predicts the velocity of the system using only \(\widetilde{x}\) and \(\widetilde{u}\) as inputs.

These results validate the use of the Steady-State PINC model as an initial condition estimator (as outlined in Section~\ref{ssec:simplifying_ic}) for training the Transient PINC model. 

\subsubsection{PINC and MPC Controller in the Transient Regime}\label{subsubsection:mpc_inc}

\begin{figure}[h!]
    \centering
    \includegraphics[width=1.0\textwidth]{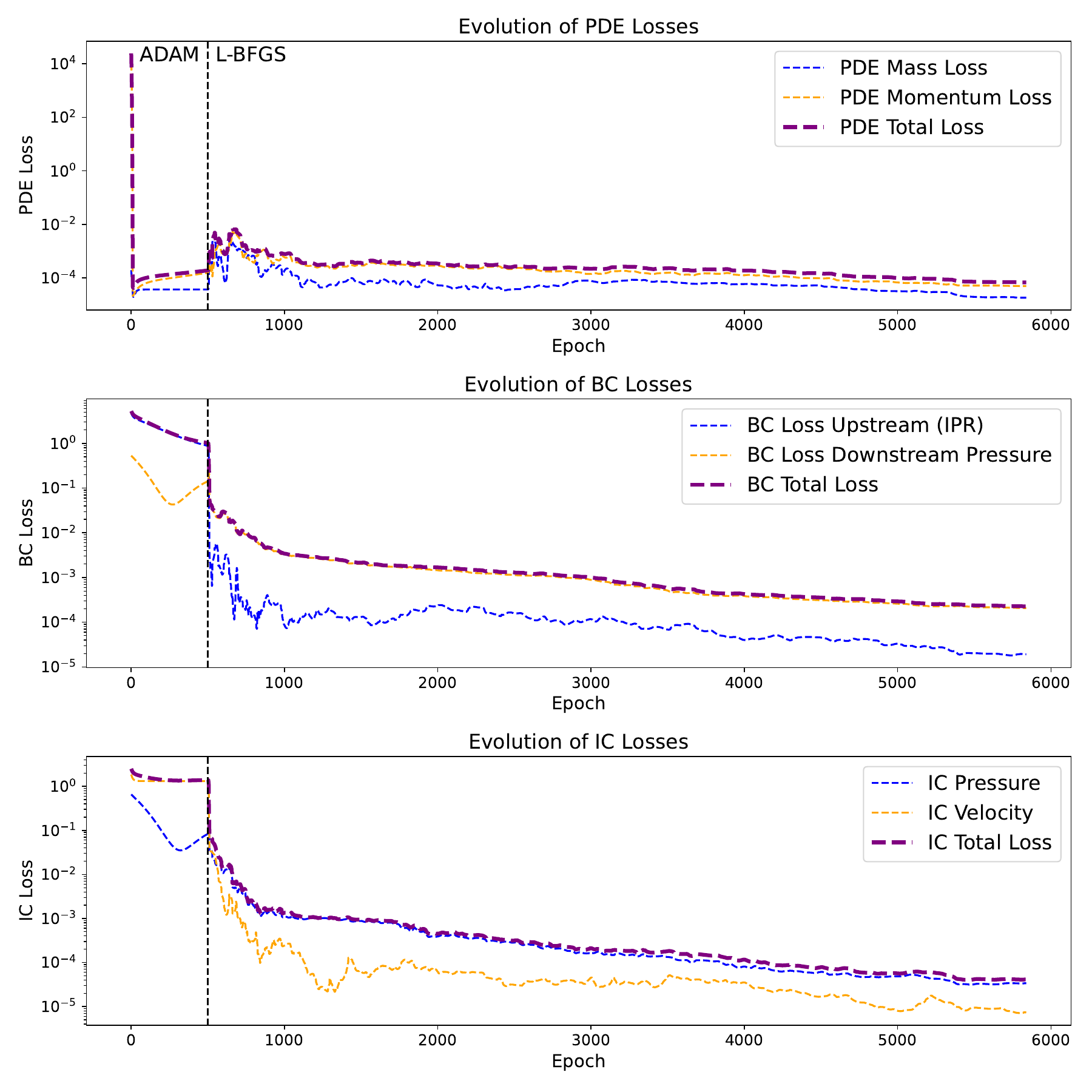}
    \caption{The training strategy begins with a coarse optimization using the ADAM optimizer for 300 epochs (indicated by the black vertical dashed line), followed by successive refinement with the L-BFGS optimizer. The L-BFGS method effectively refines the solution, as evidenced by the significant reduction in both the boundary condition (BC) and initial condition (IC) losses during the optimizer transition, maintaining consistency until convergence. Numerical compatibility due to normalization (Section \ref{subsubsection:normalized_inc}) is observed between the losses for the mass and momentum conservation equations and those for the boundary and initial conditions. The validation error decreases consistently as training progresses, indicating the absence of overfitting. The loss magnitudes are on the order of \(10^{-5}\), demonstrating that the model has good representativeness.}
    \label{fig:Transient_training}
\end{figure}

For the transient PINC model described in Section~\ref{sec:pinc_t}, we employed a fully connected neural network (MLP) with 4 hidden layers, each containing 20 neurons. The hyperbolic tangent (\(\tanh\)) activation function was used across all layers of the model. Optimization was carried out in two stages: an initial phase of 300 iterations with the ADAM optimizer to initialize the parameters, followed by fine-tuning with the L-BFGS optimizer, as depicted in Figure \ref{fig:Transient_training}. Unlike the PINC steady-state model (described in section \ref{sec:pinc_ss}), the transient model incorporates 4 input features, including two additional inputs representing time and initial conditions.

The total number of collocation points used for the PDE loss is \(N_{\mathcal{F}} = 10000\). For the boundary condition loss, \(N_{\mathcal{B}} = 2000\) points were utilized, evenly distributed between the two boundaries: the upstream boundary, where the IPR equation is imposed, and the downstream boundary, where the pressure is specified. For the initial condition loss, \(N_{\mathcal{I}} = 1000\) points were utilized calculated from the steady-state assuming their weights are fixed. Since the model predicts two outputs (pressure and velocity), the number of outputs is \(N_y = 2\).

The transient PINC model offers significant potential for monitoring applications, providing detailed insights into the spatiotemporal evolution of profiles as the dynamic system unfolds. Beyond monitoring, its utility extends to optimization and control, which constitutes the central focus of this work. Specifically, we aim to rigorously evaluate the model's predictive capabilities in forecasting the system's behavior when subjected to a predefined control signal trajectory, \(\widetilde{u}_1, \widetilde{u}_2, \ldots, \widetilde{u}_{k}\).

The control signal trajectory is derived from the model predictive control (MPC) framework described in Section~\ref{subsubsection:mpc_Transient}. In this application, it is worthwhile mentioning some practical aspects:
\begin{itemize}
    \item $\bar{x}$ represents a fixed position where sensor measurements are available. In oil and gas systems, this typically corresponds to the location of a permanent downhole gauge (PDG), which is often installed near the bottom of the well. In our application, the variable of interest, $y_{\mathrm{target}}$, corresponds to the target pressure measured at $\bar{x}$.
    
    \item The selection of $y_{\mathrm{target}}$ is driven by the objective of maximizing mass flow production in the well. This is based on the IPR Equation (\ref{eq:IPR_condition}), which suggests that maximizing the pressure drawdown, defined as $P_{reservoir} - P(x=0, t)$, leads to increased flow rates. To achieve this, $y_{\mathrm{target}}$ is set to a low value, such as zero, adopting the strategy of an unfeasible target to drive production to its maximum potential.
    
    \item Dynamical constraints are imposed to limit the maximum pressure variation at the PDG, denoted as $\Delta y_{\mathrm{max}}$. These constraints ensure that the pressure variation remains within allowable limits over the sampling period $\widetilde{T}_s$, introducing smoothness into the control sequence. This practical consideration prevents abrupt changes in the manipulated variables controlling the plant; otherwise, the optimizer's solution would likely be a sudden step change to minimize the objective function.
    
    \item The prediction horizon, $N_p$, is set to 10, while the control horizon, $N_u$, is set to 2. At each sampling time, feedback is collected from the plant, and the measured signal is used to update the MPC prediction. A smaller control horizon is chosen to simplify the optimization problem described in Section~\ref{subsubsection:mpc_Transient}, ensuring computational efficiency.

\end{itemize}

Before explicitly demonstrating the effect of the MPC controller, we first aim to evaluate the performance of the transient PINC model under a predefined trajectory of manipulated variables, \(\widetilde{u}_1, \widetilde{u}_2, \ldots, \widetilde{u}_k\), generated by the MPC controller. Specifically, our objective is to assess the model's behavior using these control signals in an open-loop simulation.

\begin{figure}[h!]
    \centering
    \includegraphics[width=1.0\textwidth]{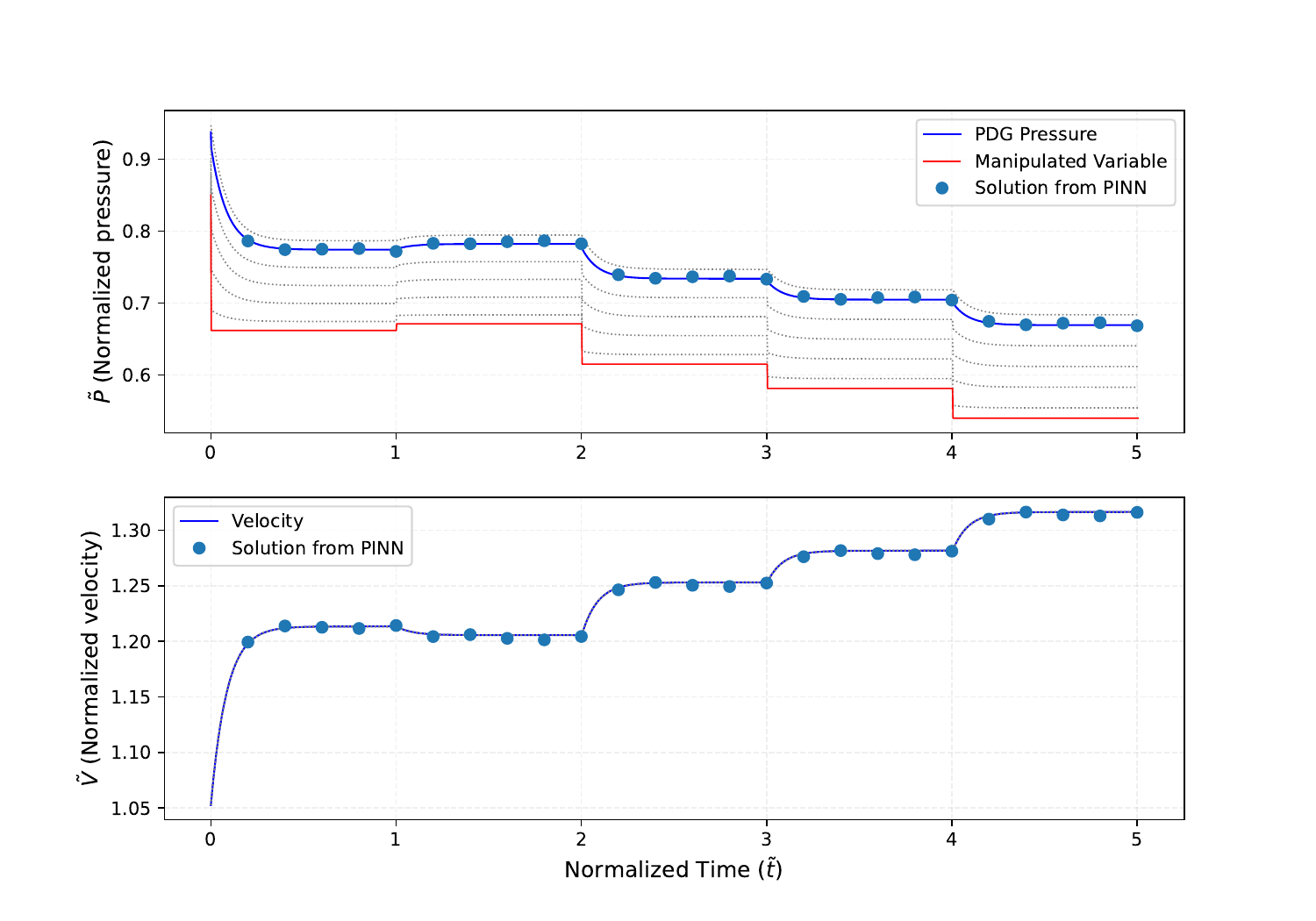}
    \caption{It is observed that the transient PINC model accurately captures the system's dynamics, particularly when the window time is sufficiently large (\textit{e.g.}, $t_{\text{ref}}$ = 10 seconds) to allow the plant to stabilize. The control trajectory is applied to the plant and is calculated from a strategy using MPC control. Here, the MPC-derived controls are applied in open loop, and we observe that both the spatiotemporal profiles of pressure and velocity calculated by the PINC are consistent with those numerically simulated. It is worth noting that, as the flow is incompressible, the velocity does not vary spatially. The dotted variables in the pressure graph (in gray) indicate how pressure varies spatially along the system. Specifically, the pressure trend highlighted in the graph corresponds to the location where instrumentation is available, at the PDG. Given that the model has proven sufficiently accurate in capturing the system's dynamics, it can be employed as a surrogate model for the Model Predictive Controller (MPC).}
    \label{fig:Transient_10sec}
\end{figure}

This evaluation is illustrated in Figure~\ref{fig:Transient_10sec}, which depicts a 10-second time window. The manipulated variable remains piecewise constant within each time window, and the results from the PINC forward simulation, as described in Algorithm \ref{alg:forward-sim}, show strong agreement with those obtained using finite difference methods, which are considered the reference plant model. The velocity spatial profile converges, as the flow is assumed to be incompressible, while the solution from the transient PINC forward simulation closely matches the observed data at the permanent downhole gauge (PDG). Notably, the pressure recorded at the PDG demonstrates excellent alignment between the PINC model and the plant. Furthermore, the entire pressure profile, represented for all positions with dotted gray lines, corresponds accurately to the predictions of the PINC model.

These results are consistent with expectations, as the 10-second time window is sufficiently large to allow the system to stabilize. This ensures that the initial condition, calculated based on the control signal from the previous time step (as discussed in Section  \ref{ssec:simplifying_ic}), aligns well with the system's dynamic evolution, making the open-loop PINC simulation accurate for this purpose. 

If the time window were smaller, larger deviations are observed between the open-loop simulation and the PINC prediction. This behavior is indeed seen in control systems with a sampling time of 1 second. However, as feedback from the plant is incorporated at each sampling time, these deviations are dynamically corrected. Even under these conditions, the PINC model, acting as the predictor in the Model Predictive Control (MPC) framework, remains sufficiently accurate for closed-loop control applications.

\begin{figure}[h!]
    \centering
    \includegraphics[width=1.0\textwidth]{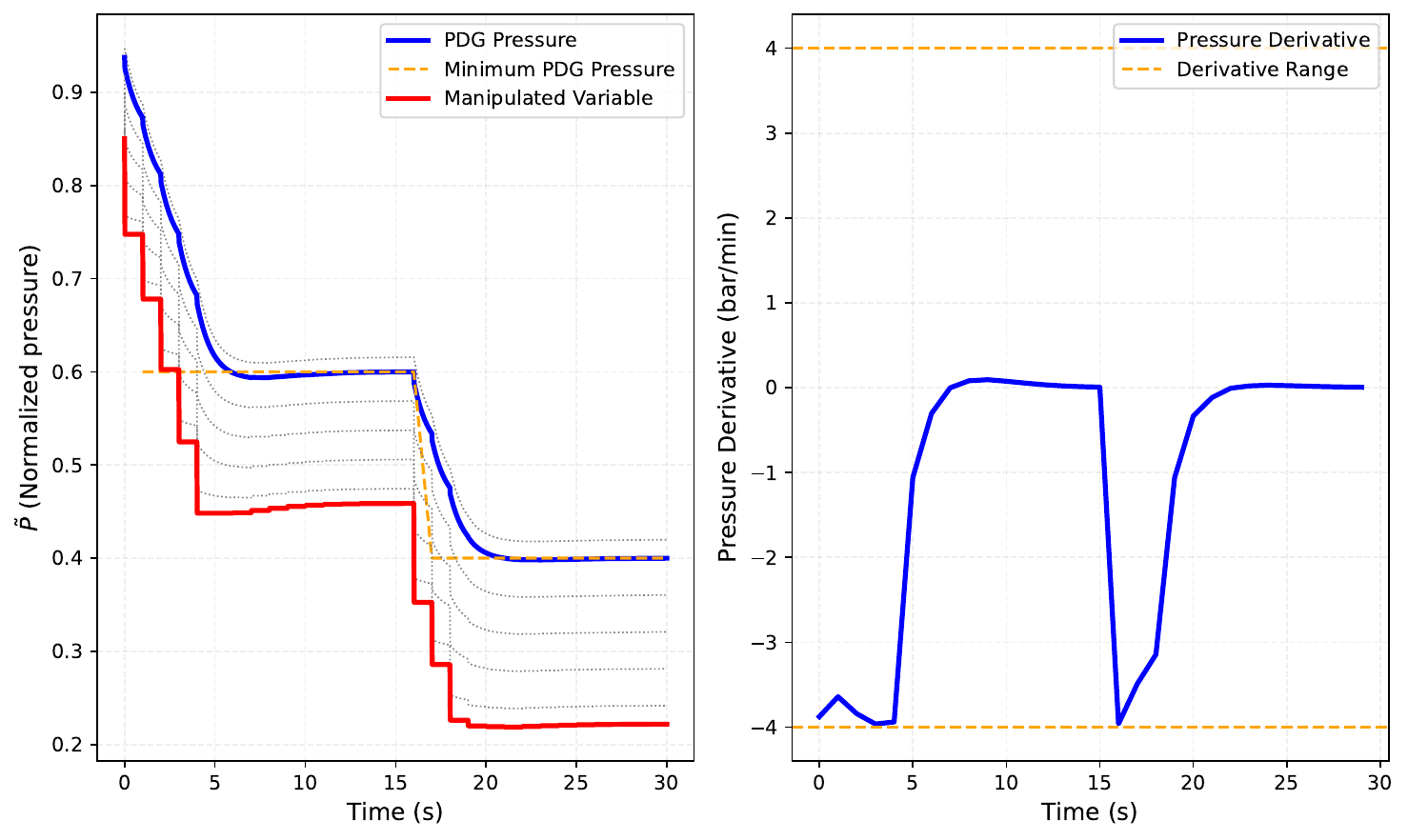} 
    \caption{Evolution of the pressure measured at the PDG (\( \widetilde{x} = 0.1 \)), where, starting from an initial condition, a change in the control trajectory (\( \widetilde{u} \)) is determined to achieve a target for the PDG pressure. The chosen target is zero, an unattainable target, which drives the maximization of production. Two constraints are activated in this process: the first concerns the pressure derivative at the PDG (limited to 4 bar/min), which prevents the controller from aggressively reaching the minimum PDG pressure. Once the minimum PDG pressure is achieved, the controller performs only small adjustments to the manipulated variable through model correction via feedback, ensuring compliance with the minimum BHP constraint. At \( t = 15 \,\mathrm{s} \), the minimum PDG pressure is further reduced, allowing the controller to decrease the manipulated variable even further, while still respecting the dynamic constraints imposed by the MPC. This ensures a gradual and safe transition towards the new operating point while respecting the operational dynamical constraints.
}
    \label{fig:mpc_feedback}
\end{figure}

The closed-loop system utilizing the MPC controller with a sampling time of 1 second is illustrated in Figure~\ref{fig:mpc_feedback}. In this setup, the manipulated variable is updated every \( T_\text{s} \). The control horizon, \( N_c \), defines the number of control variables predicted over a future time window. However, at each sampling step, only the first control action, \( \widetilde{u}_{1}\), is applied to the plant. This process is iteratively repeated in the closed-loop system, enabling the controller to leverage measurement signals, update its predictions dynamically, and utilize the PINC model's forecasts to derive the control sequence effectively.

Naturally, the system tends to maximize production by reducing the bottom-hole pressure (BHP) as quickly as possible. However, the imposed constraint on the rate of pressure variation at the PDG, limited to 4 bar/min, ensures that this reduction is achieved with a smoother dynamic response. This balance between aggressive production optimization and adherence to operational constraints is made possible by the predictive capabilities of the transient PINC model. By accurately forecasting the system's behavior, the model empowers the MPC controller to navigate complex dynamics effectively, handling operational constraints while driving the system towards its optimal performance. This showcases the robust potential of this approach for advanced production optimization.

\clearpage

\subsection{Compressible Flow}

This section addresses the flow problem under the assumptions of isothermal and compressible flow. The case study considers a long pipeline transporting gas, as opposed to the water flow discussed in the incompressible flow scenario. 

The parameters used in the simulation of the compressible single-phase gas flow system are summarized in Table~\ref{tab:simulation_parameters_compressible}. The pipe diameter (\(D\)) is set to 0.2 meters, and the fluid viscosity (\(\mu\)) is \(5 \times 10^{-5}\) Pa·s. The IPR parameter (\(\text{PI}\)) is chosen as \(5 \times 10^{-4} \frac{\text{kg}}{\text{s} \cdot \text{Pa}}\). The table also presents reference values for pressure, velocity and density, which are used in the normalization of balance and momentum governing equations, as described in Section \ref{subsubsection:normalized_comp}.

The static pressure (\(P_{\text{reservoir}}\)) at the upstream boundary is set to \(50 \times 10^5\) Pa, and the total length of the pipe is 2000 meters. The temperature is assumed to be constant at 300 K.

\begin{table}[h!]
    \centering
        \caption{Simulation parameters used for the compressible and horizontal single-phase gas flow system, including MPC controller settings (3 last rows).}
    \label{tab:simulation_parameters_compressible}
    \begin{tabular}{|l|c|c|}
        \hline
        \textbf{Parameter} & \textbf{Symbol} & \textbf{Value} \\ \hline
        Pipe Diameter & \(D\) & 0.2 m \\ \hline
        Fluid Viscosity & \(\mu\) & \(5 \times 10^{-5}\) Pa·s \\ \hline
        IPR Parameter & \(\text{PI}\) & \(5 \times 10^{-4} \, \frac{\text{kg}}{\text{s} \cdot \text{Pa}}\) \\ \hline
        Reservoir Pressure & \(P_{\text{reservoir}}\) & \(50 \times 10^5\) Pa \\ \hline
        Total Pipe Length & - & 2000 m \\ \hline
        Inclination & \(\theta\) & 0 \\ \hline
        Pressure Reference & \(P_{\text{ref}}\) & \(50 \times 10^5\) Pa \\ \hline
        Velocity Reference & \(V_{\text{ref}}\) & 50 m/s \\ \hline
        Density Reference & \(\rho_{\text{ref}}\) & 60 kg/m\(^3\) \\ \hline
        Time Reference & \(t_{\text{ref}}\) & 100 s \\ \hline
        Friction Factor Calculation & \(f\) & Swamee-Jain \\ \hline       
        Control Horizon & \(N_c\) & 2 \\ \hline
        Prediction Horizon & \(N_p\) & 10 \\ \hline
        Sampling Time & \(T_s\) & 10 s \\ \hline        
    \end{tabular}
\end{table}

\subsubsection{Model Selection with Optuna}\label{subsubsection:optuna}

The modeling of compressible flow systems is significantly more complex than that of incompressible systems. Using an arbitrary number of neurons in a simple feedforward neural network architecture (as described in Section \ref{subsection:inc_flow_results} for the incompressible case) is not sufficient to achieve the desired results. This complexity arises from the mass and momentum conservation equations, which do not adhere to the simplifying assumptions used in incompressible systems (as described in Section \ref{subsection:compressible_gas_flow}), and from an equation of state that considers the spatiotemporal dependence of density, resulting in a significantly more complex behavior. Furthermore, as gas flow systems are highly compressible, transient responses, particularly step-type inputs, can produce highly oscillatory outputs, further complicating the training of neural networks, which tend to smooth the system's responses.

To address these challenges, we perform model selection by varying several neural network hyperparameters using Optuna. Optuna is an open-source framework for hyperparameter optimization that automates the search process through efficient sampling techniques, such as Tree-structured Parzen Estimators (TPE) and multi-armed bandit strategies, to identify optimal configurations within a predefined search space \citep{akiba2019optuna}. By employing Optuna, the hyperparameter tuning process becomes more efficient and less reliant on manual trial-and-error approaches, thus improving the model's performance and convergence speed. The hyperparameters of the neural network model are listed in Table \ref{tab:hyperparameters}.

The model selection strategy is computationally expensive, as for each hyperparameter combination, we perform 5 training runs with 5 predefined seeds (for both the sampling points via Latin Hypercube Sampling (LHS) and the neural network's weights) and take the median of the validation loss. This process becomes less costly as the number of iterations of the L-BFGS algorithm is limited. However, it still results in a significant computational burden.

Therefore, we use Optuna as a guide to provide, after 100 iterations, a suitable and robust selection of hyperparameters for model training. Moreover, to choose the final model, we select the set of hyperparameters corresponding to the best seed with the lowest validation loss throughout the Optuna search process and further refine the results by increasing the maximum number of iterations of the L-BFGS algorithm, allowing it to converge naturally and stop the search process on its own.

Some hyperparameters were not explicitly defined throughout this text. Therefore, we provide a detailed explanation below:
\begin{itemize}
    \item Sinusoidal Activation Function:  
    The sinusoidal activation function is expressed as:
    \[
    f(x) = w_1 \sin(x) + w_2 \cos(x)
    \]
    where \(w_1\) and \(w_2\) are trainable weights for each layer. This activation function is particularly useful for learning periodic patterns, which are common in physical systems described by differential equations.

    \item Swish Activation Function:  
    The Swish (or SiLU) function is defined as:
    \[
    \text{Swish}(x) = \frac{x}{1 + e^{-x}}
    \]
    This function was introduced by \cite{ramachandran2017searching} and has been shown to improve performance in deep networks by providing smooth and non-monotonic activation, which helps with better gradient flow compared to traditional functions like ReLU.

    \item Skip Connections:  
    Skip connections \citep{ANTONELO2024127419, he2016deep,wang2021understanding} are structures that introduce direct connections between the input layer and intermediate layers in the network. In addition to the fully connected dense network, these connections leverage additional layers, referred to as encoders, which take the input directly and use it to influence the activation of each layer, except the final one. This architecture ensures that each layer maintains a strong relationship with the input layer, thereby improving gradient flow and mitigating the vanishing gradient problem observed in deep models.

    This structure implements a neural network with $N_L$ hidden layers of $N_n$ neurons each. The input $X$ is projected into a higher-dimensional space through the encoder layers by using their respective weights and biases $W^1, W^2, b^1,$ and $b^2$. The transformations for $U$ and $V$ are given by:
    \begin{equation} \label{eq:UV}
     \left \{ \begin{aligned}
    U &= \phi(W^1 X + b^1)      \\
    V &= \phi(W^2 X + b^2) 
    \end{aligned} \right .
    \end{equation}
    where $\phi$ is the activation function.
        
    The first step for calculating the forward pass is a standard pass through a dense layer to compute $Z^{(1)}$, as shown in Equation (\ref{eq:z1}):
    \begin{equation}
    Z^{(1)} = \phi(W^{z,1} X + b^{z,1})    \label{eq:z1}
    \end{equation}
    
    Then this $Z^{(1)}$ is weighted by element-wise multiplication ($\odot$) with the encoder vectors $U$ and $V$ to calculate the activation $A^{(1)}$ of the first layer:
    \begin{equation}
    A^{(1)} = (1 - Z^{(1)}) \odot U + Z^{(1)} \odot V      \label{eq:a1}
    \end{equation}
    
    This propagation continues through all the remaining $N_L$ layers:
        \begin{equation} \label{eq:zk-ak} 
    \left \{ \begin{aligned}
    Z^{(k)} &= \phi(W^{z,k} A^{(k-1)} + b^{k})    \\
    A^{(k)} &= (1 - Z^{(k)}) \odot U + Z^{(k)} \odot V, \quad k = 2, \dots, N_L
    \end{aligned} \right .
    \end{equation}
    The final output $y$ is calculated as a linear projection of the previous layer’s activation without the weighting from the encoder layers:
    \begin{equation}
    y = W^{out} A^{(N_L)} + b^{out}    \label{eq:y}
    \end{equation}
    
\end{itemize}

\begin{table}[H]
\centering
\caption{Optimized Hyperparameters for Steady-State and Transient PINC Models.}
\resizebox{\textwidth}{!}{%
\begin{tabular}{|l|cc|cc|}
\hline
\textbf{Hyperparameter}       & \multicolumn{2}{c|}{\textbf{Steady-State}} & \multicolumn{2}{c|}{\textbf{Transient}} \\ \hline
                              & \textbf{Range}     & \cellcolor{green!20}\textbf{Solution}  & \textbf{Range}      & \cellcolor{green!20}\textbf{Solution}  \\ \hline
\# of Layers                  & 3 -- 8             & \cellcolor{green!20}8                  & 3 -- 8              & \cellcolor{green!20}8                 \\ \hline
Hidden Size                   & 10 -- 100          & \cellcolor{green!20}43                 & 10 -- 100           & \cellcolor{green!20}93                \\ \hline
Activation Function           & \{\texttt{tanh, sinusoidal}\} & \cellcolor{green!20}tanh               & \{\texttt{tanh, sinusoidal, swish}\} & \cellcolor{green!20}swish         \\ \hline
\# of Collocation Points (\(N_{\mathcal{F}}\))      & 500 -- 10000       & \cellcolor{green!20}8723               & 500 -- 10000        & \cellcolor{green!20}4608              \\ \hline
\# of BC Points (\(N_{\mathcal{B}}\))              & 50 -- 1000         & \cellcolor{green!20}434                & 50 -- 2000          & \cellcolor{green!20}1449              \\ \hline
\# of IC Points (\(N_{\mathcal{I}}\))              & --                 & \cellcolor{green!20}--                 & 50 -- 2000          & \cellcolor{green!20}1213              \\ \hline
\# of Epochs (ADAM)           & 500 -- 1500        & \cellcolor{green!20}1095               & 300 -- 700          & \cellcolor{green!20}699               \\ \hline
Skip Connections              & \{\texttt{true, false}\}    & \cellcolor{green!20}false              & \{\texttt{true, false}\}     & \cellcolor{green!20}true             \\ \hline
\end{tabular}%
}
\label{tab:hyperparameters}
\end{table}

The optimized hyperparameters for the steady-state and transient models presented in Table \ref{tab:hyperparameters} provide valuable insights into the behavior of physics-informed neural networks (PINNs) applied to compressible flow problems. The selection process, conducted using Optuna, highlights several key characteristics of these models.

Optuna chose a relatively high number of hidden layers for both steady-state and transient models, suggesting that deep multi-layer perceptrons (MLPs) tend to generate more representative models of complex physical phenomena, such as compressible flow in pipelines. This indicates that deeper networks are capable of capturing intricate spatial and temporal relationships inherent in fluid flow systems.

However, the number of neurons per layer differed significantly between the two cases. The transient model required 93 neurons per layer, whereas the steady-state model only required 43 neurons per layer. This result emphasizes the fact that transient models deal with higher levels of complexity due to the inclusion of time-dependent dynamics. In contrast, steady-state models focus solely on the spatial distribution of state variables, which results in a less complex problem.

Regarding the choice of activation function, the swish function was identified as the most suitable for the transient model, while tanh was chosen for the steady-state case. The swish function has been shown to outperform traditional activation functions such as ReLU and tanh in capturing nonlinear interactions. This is particularly relevant for transient flow problems, where variables like pressure and velocity exhibit complex, oscillatory behavior and intricate temporal dynamics. Additionally, skip connections are necessary in the compressible transient case to improve gradient flow and training stability.

Another important observation is the distribution of collocation points and boundary/initial condition points. In both models, the number of collocation points (\(N_{\mathcal{F}}\)) significantly exceeded the number of boundary/initial condition points (\(N_{\mathcal{B}}\) and \(N_{\mathcal{I}}\)). This behavior is consistent with the findings of \cite{RAISSI2019686}, who demonstrated that PINNs tend to perform better when more collocation points are used to enforce the underlying partial differential equations (PDEs). The larger number of collocation points ensures that the solution adheres closely to the governing equations throughout the domain.

The number of epochs used for pre-training with the ADAM optimizer is another critical hyperparameter. The steady-state model required 1095 epochs, whereas the transient model used 699 epochs. This pre-training phase is crucial because it places the model in a reasonable parameter space before switching to the L-BFGS optimizer for fine-tuning. A longer ADAM pre-training phase can lead to poor convergence during the L-BFGS phase, as the model may become stuck in unfavorable regions of the loss landscape. Conversely, too few epochs can leave the model underprepared, making it difficult for L-BFGS to complete the training effectively. Therefore, the balance between these two phases represents a trade-off that requires careful consideration. The chosen number of epochs in this case reflects an attempt to achieve this balance.

\subsubsection{PINC in the Steady-State Regime} \label{subsubsection:pinc_ss_compressible}

For the steady-state PINC model in compressible flow systems described in Sections~\ref{sec:pinc_ss} and \ref{subsection:compressible_gas_flow}, we presented in the previous section the hyperparameter selection and the chosen neural network architecture. In this section, we will briefly discuss the underlying physics of compressible flow in the Steady-State Regime and how the neural network captures the spatial profiles of state variables, such as pressure and velocity.

\begin{figure}[h!]
    \centering
    \includegraphics[width=1.0\textwidth]{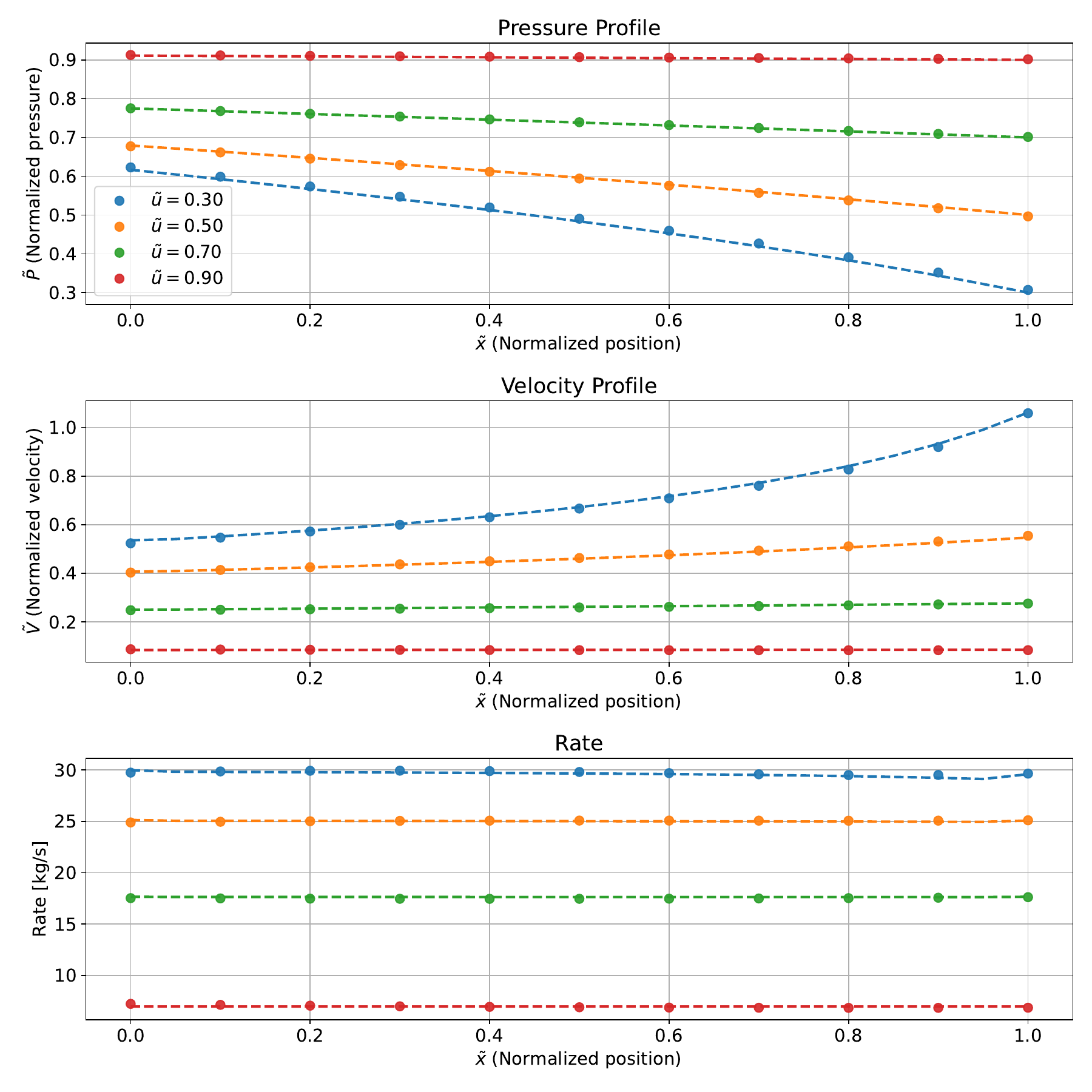}
    \caption{Comparison between numerically simulated results (dashed line) and those computed by the PINC model (solid points). The Steady-State PINC performs as expected, capturing the spatial trend of pressure and velocity distributions across a range of control values ($\widetilde{u}$). Note that, unlike incompressible flow behavior, the velocity is not constant along the x-axis. In the steady-state regime, due to mass conservation (Equation \ref{eq:mass}), it is the mass flow rate that remains constant, as seen in the third subplot. Since pressure decreases along the flow, density also decreases, forcing velocity to increase to maintain the mass flow. This effect is particularly noticeable for lower values of $\widetilde{u}$. These results support the use of the Steady-State PINC model as an initial condition estimator (as outlined in Section \ref{ssec:simplifying_ic}) for training the Transient PINC model.}

    \label{fig:simulation_pinc_ss}
\end{figure}

In steady-state flow, the mass flow rate is conserved along the flow axis. Equation \eqref{eq:mass_conservation} for mass conservation implies that $\frac{d(\rho V)}{dx} = 0$ for the steady-state regime. Expanding this to include the constant cross-sectional area $A$, we have $\frac{d(\rho A V)}{dx} = 0$. Since $A$ is constant in this case, the equation reduces to $\frac{d\dot{m}}{dx} = 0$, where $\dot{m}$ is the mass flow rate (as defined in Equation \ref{eq:mass_rate_definition}). Thus, in steady-state conditions, the mass flow rate remains constant along the flow direction. Therefore, for each downstream boundary condition where the pressure ($\widetilde{u}$) is known, there is a corresponding mass flow rate, as shown in the third subplot of Figure \ref{fig:simulation_pinc_ss}.

The second key aspect of this analysis involves the behavior of the state variables: pressure, density, and velocity. As shown in Figure \ref{fig:simulation_pinc_ss}, there is a pressure drop along the flow due to frictional losses. Consequently, the pressure decreases along the flow direction ($\widetilde{x}$). Since the density is proportional to pressure (Equation \ref{eq:eos_ideal_gas}), the density also decreases along the flow.

Given that the mass flow rate is constant ($\dot{m} = \rho A V$), the velocity must increase along $x$ to compensate for the decrease in density.  This rise in velocity reflects the fluid's acceleration, which is significant in this scenario, contrasting with the negligible acceleration observed in incompressible flow.

\newpage

\subsubsection{PINC and MPC Controller in the Transient Regime}\label{subsubsection:pinc_transient_compressible}

For the Transient PINC model in compressible flow systems described in Sections~\ref{sec:pinc_t} and \ref{subsection:compressible_gas_flow}, we  presented  above (Section \ref{subsubsection:optuna}) the hyperparameter selection and the chosen neural network architecture. In this section, we will briefly discuss the underlying physics of compressible flow in the Transient Regime and how the neural network captures the spatiotemporal profiles of state variables, such as pressure and velocity.

\begin{figure}[h!]
    \centering
    \includegraphics[width=1.0\textwidth]{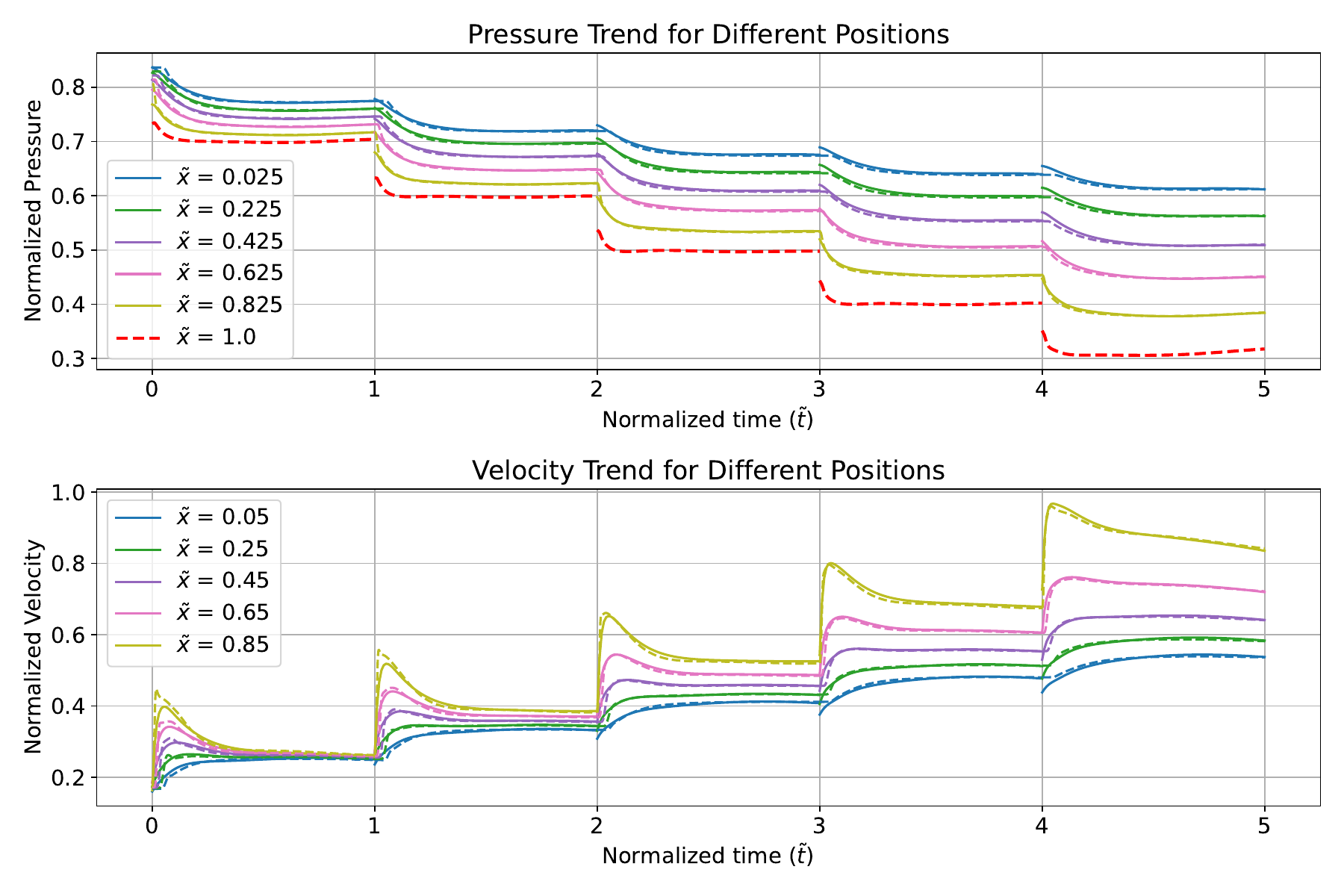}
    \caption{Comparison between the transient PINC solution (solid lines) and the numerical solution (dashed lines) for compressible flow. The first subplot highlights the pressure profile at different points ($\widetilde{x}$), with the dashed red line representing the neural network's output applied to the plant as a downstream boundary condition. The second subplot shows the velocity profile, where a transient effect, characterized by an increase in velocity, is observed as the valve is gradually opened.}

    \label{fig:Transient_comparison_compressible}
\end{figure}

Figure \ref{fig:Transient_comparison_compressible} compares the Transient PINC solution, represented by solid lines, with the numerical solution, represented by dashed lines. Five time windows within the interval $[0, t_{\text{ref}}]$ were considered, each corresponding to a $\widetilde{u}$ value (the neural network's input) of 0.7, 0.6, 0.5, 0.4, and 0.3, respectively.
The transient PINC solution was obtained using the Forward Simulation method detailed in Section \ref{subsubsection:forward_simulation}.

This simulation represents a gradual valve opening, where the flow stabilizes upon reaching the steady state, followed by subsequent changes in control inputs. The region influenced by the choice of $\widetilde{u}$ in this simulation exhibits greater complexity due to phenomena such as fluid acceleration.

The first subplot of Figure \ref{fig:Transient_comparison_compressible} shows the pressure trend at different points along the flow ($\widetilde{x}$). In this subplot, the PINC output at $\widetilde{x} = 1$ is highlighted in red dashed lines. Ideally, if the boundary condition loss for the upstream were zero, we would have perfect step changes instead of smoother transitions. However, since the BC loss is part of the objective function, smoother transitions between control windows are observed. This smoothness in modeling dynamic systems subject to boundary conditions (or controls) is an interesting property of PINNs, as their inherent modeling self-regulates, avoiding highly oscillatory or overly rapid behaviors that might arise with hard step constraints.

The second subplot highlights a more complex variable, focusing on the velocity trend. A gradual opening of the choke is presented, where greater spatial dispersion in the velocity profile is observed as $\widetilde{u}$ decreases. For example, the velocity spatial dispersion for $\widetilde{u} = 0.7$ (first window) is smaller compared to $\widetilde{u} = 0.3$ (last window). It is observed that the velocity peak becomes more pronounced for $\widetilde{x}$ values closer to 1, where the boundary condition is imposed as a control signal.

In upstream sections of the flow (as $\widetilde{x}$ approaches 0), the velocity exhibits a more stable profile, constrained by the upstream boundary condition of the IPR type (Equation \ref{eq:IPR_condition_mass}). This behavior is physically consistent, as the mass source  adjusts its input to the system in response to the smooth temporal variations in the pressure profile.

The comparison points ($\widetilde{x}$) for the pressure and velocity profiles are different. While this is irrelevant for the PINC model, which can compute outputs at any position $\widetilde{x}$, the numerical scheme used for comparison is based on cells. Pressure and density cells are defined at their centers, while velocity is defined at the boundaries. Therefore, to perform this comparison with the PINC, there is a positional offset between the pressure and velocity variables.

The control signal trajectory is derived from the model predictive control (MPC) framework described in Section~\ref{subsubsection:mpc_Transient}. For the compressible system, it is worthwhile mentioning some practical aspects:

\begin{itemize}
    \item The selection of $y_{\mathrm{target}}$ is guided by the objective of maximizing mass flow production. To achieve this, $y_{\mathrm{target}}$ is set to a low value—often zero—following an unattainable target for the PDG pressure that drives production to its maximum potential.
    
    \item Dynamical constraints are introduced to limit both the maximum pressure variation at the PDG and the manipulated variable. Unlike the original formulation (\ref{eq:mpc_transient}), where the manipulated variable is treated as a soft constraint with a penalty in the objective function, here it is imposed as a hard constraint to explore different controller behaviors.
    
    \item The prediction horizon, $N_p$, is set to 10, while the control horizon, $N_u$, is set to 2. At each sampling time, feedback is collected from the plant, and the measured signal is used to update the MPC prediction.

\end{itemize}

\begin{figure}[h!]
    \centering
    \includegraphics[width=1.0\textwidth]{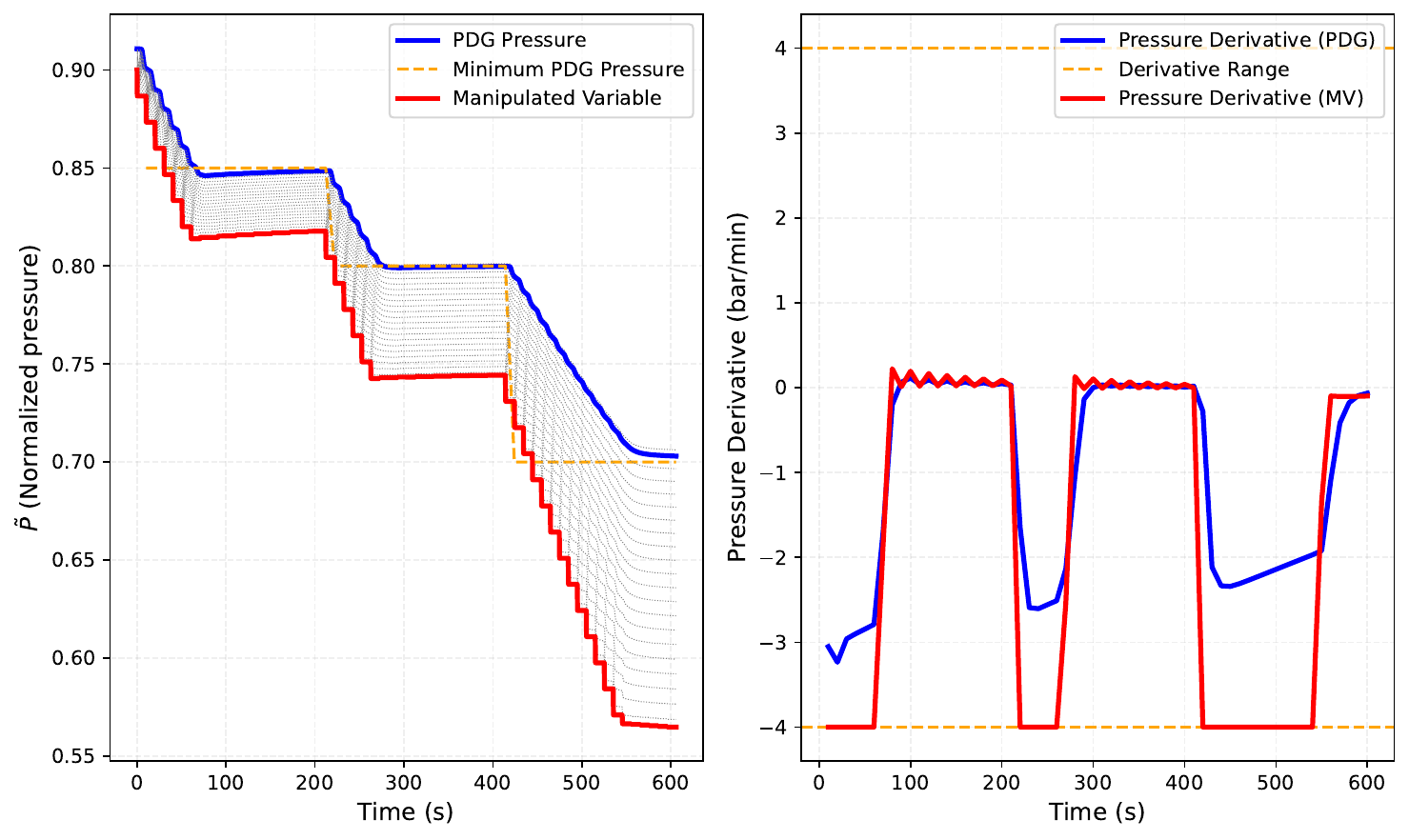} 
    \caption{Evolution of the pressure measured at the PDG (\( \widetilde{x} = 0.075 \)), where, starting from an initial condition, a change in the control trajectory (\( \widetilde{u} \)) is determined to achieve a target for the PDG pressure. The chosen target is zero, an unattainable target, which drives the maximization of production. Two constraints are activated in this process: the first concerns the pressure derivative for the manipulated variable (limited to 4 bar/min, indicated in red), which prevents the controller from aggressively reaching the minimum PDG pressure. Once the minimum PDG pressure is achieved, the controller performs only small adjustments to the manipulated variable through model correction via feedback, ensuring compliance with the minimum PDG pressure constraint. At \( t = 200 \,\mathrm{s} \) and \( t = 400 \,\mathrm{s} \), the minimum PDG pressure is further reduced, allowing the controller to decrease the manipulated variable even further, while still respecting the dynamic constraints imposed by the MPC. This plot demonstrates the suitability of the PINC model for MPC applications, where the controller successfully generates control signals directly applied to the plant across a wide range of operating points. This capability is particularly noteworthy, as the system is highly nonlinear, yet the model maintains its forecasting accuracy without requiring retraining or adaptation.
}
    \label{fig:mpc_feedback_compressible}
\end{figure}

The closed-loop system utilizing the MPC controller with a sampling time of 10 seconds (\( T_\text{s} \)) is illustrated in Figure~\ref{fig:mpc_feedback_compressible}. In this setup, the manipulated variable is updated every \( T_\text{s} \) seconds. The control horizon, \( N_c \), defines the number of control variables predicted over a future time window. However, at each sampling step, only the first control action, \( \widetilde{u}_{1}\), is applied to the plant. This process is iteratively repeated in the closed-loop system, enabling the controller to leverage measurement signals, update its predictions dynamically, and utilize the PINC model's forecasts to derive the control sequence effectively.

Naturally, the system tends to maximize production by reducing the bottom-hole pressure (BHP) as quickly as possible. However, the imposed constraint on the rate of pressure variation at the PDG and for the manipulated variable, limited to 4 bar/min, ensures that this reduction is achieved with a smoother dynamic response.

\clearpage
\subsection{Model Assessment}\label{model_assessment}

This work presents results for both compressible and incompressible flows. For each case, a model was developed using physics-informed neural networks for control (PINC) to approximate the solutions under steady-state and transient regimes. This section aims to compare and quantify the results in terms of error metrics and computational execution time.

To perform this comparison, the numerical solution obtained via finite differences is used as the ground truth. This numerical solution relies on a spatial and temporal grid, composed of cells representing spatial discretization points where the variables are defined. The temporal grid is defined as the solution advances through time with discrete time steps. Consequently, the comparison requires using the same spatiotemporal cells from the numerical solution for the PINCs.

For the steady-state regime, we compare the numerical solution with the steady-state PINN (PINC-SS) and with the transient PINC at the initial time (PINC-Transient at $\tilde{t}=0$). The comparison focuses on the steady-state solution for different positions depending on the applied control. The solution for each pair $(\tilde{x},\tilde{u})$ is a scalar, for both velocity and pressure, and an appropriate metric to quantify the model's accuracy is the Mean Absolute Percentage Error (MAPE), defined as:
\begin{equation*}
   \text{MAPE} = \frac{1}{N}\sum_{i=1}^{N} \left|\frac{y_{\text{true}, i} - y_{\text{est}, i}}{y_{\text{true}, i}}\right| \times 100\%
\end{equation*}
where $y_{\text{true}}$ and $y_{\text{est}}$ represent the true (from numerical solution) and predicted time series (from the PINC solution), respectively.

The obtained MAPE values (Table \ref{tab:steady_state_resume}) are sufficiently accurate (below 1.1\%) for the PINC-SS model, as calculated from the data presented in Figures \ref{fig:simulation_pinc_ss:incompressible} and \ref{fig:simulation_pinc_ss}.
It is worth noting that the compressible regime required an exhaustive hyperparameter search using Optuna to achieve these results. In contrast, the incompressible regime did not require such an effort due to its lower complexity.

When comparing the transient PINC at the initial time with the PINC-SS, we observe that transferring the initial condition from the PINC-SS model to the PINC-Transient model (during training) results in slightly higher errors when it comes to the MAPE indicator (\textit{e.g.}, \(1.5\% > 0.5\%\) for pressure).
This occurs because the initial condition is only one component of the loss function of the PINC-Transient, which also includes other terms. Nevertheless, the obtained MAPE values are relatively low (below 3.9\%), validating one of the core ideas of this work: the use of predicted initial conditions (IC) from the steady-state PINC solution in the training of the transient PINC.

\begin{table}[h]
\centering
\caption{Steady-State  Results: Execution Time Comparison and MAPE Indicator}
\renewcommand{\arraystretch}{1.3}
\setlength{\tabcolsep}{6pt}
\begin{tabular}{ccc}
\hline
\textbf{} & \textbf{Incompressible} & \textbf{Compressible} \\
\hline
PINC-SS & \cellcolor[HTML]{CAEDFB}416\textmu s ± 4.5 \textmu s  & \cellcolor[HTML]{CAEDFB}605 \textmu s ± 4.37 \textmu s  \\
Num. Solution & \cellcolor[HTML]{CAEDFB}3.17 ms ± 32.9 \textmu s  & \cellcolor[HTML]{CAEDFB}397 ms ± 2.69 ms  \\
Time Ratio & \cellcolor[HTML]{CAEDFB}7.6 & \cellcolor[HTML]{CAEDFB}656.2 \\
\hline
Pressure (PINC-SS) & \cellcolor[HTML]{DAF2D0}0.04\% & \cellcolor[HTML]{DAF2D0}0.53\% \\
Velocity (PINC-SS) & \cellcolor[HTML]{DAF2D0}0.02\% & \cellcolor[HTML]{DAF2D0}1.14\% \\
\hline
Pressure (PINC-Transient) & \cellcolor[HTML]{DAF2D0}0.99\% & \cellcolor[HTML]{DAF2D0}1.51\% \\
Velocity (PINC-Transient) & \cellcolor[HTML]{DAF2D0}0.13\% & \cellcolor[HTML]{DAF2D0}3.85\% \\
\hline
\end{tabular}
\label{tab:steady_state_resume}
\end{table}

For the transient regime, we use a time-series metric to evaluate the similarity between the predicted and ground-truth time series. The chosen metric is the Fit Compare index, defined as follows:
\begin{equation*}
  \text{Fit Compare} = \left(1 - \frac{\sqrt{\sum_{i=1}^{N}(y_{\text{true}, i} - y_{\text{est}, i})^2}}{\sqrt{\sum_{i=1}^{N}(y_{\text{true}, i} - \overline{y}_{\text{true}})^2}}\right) \times 100\%
\end{equation*}
where $y_{\text{true}}$ and $y_{\text{est}}$ represent the true (from numerical solution) and predicted time series (from the PINC solution), respectively, and $\overline{y}_{\text{true}}$ is the mean of the true time series.

Figures \ref{fig:Transient_10sec} and \ref{fig:Transient_comparison_compressible} illustrate how the transient PINC model performs at different spatial positions. For the incompressible case, we highlight the position of the pressure differential gauge (PDG), whereas multiple positions are presented for the compressible case. For each position, we obtain a time series, and Table \ref{tab:transient_resume} presents the mean fit compare index for all positions.

The results demonstrate that the metrics are satisfactory for both models, with all fit compare values exceeding 93\%. The compressible regime presents a more complex velocity profile, often exhibiting oscillatory behavior, making it challenging to identify a suitable neural network architecture. Achieving a high representativity (\textit{e.g.}, 93.9\%) required extensive hyperparameter tuning with Optuna.

\begin{table}[h]
\centering
\caption{Transient Results: Execution Time Comparison and Fit Compare index}
\renewcommand{\arraystretch}{1.3}
\setlength{\tabcolsep}{6pt}
\begin{tabular}{ccc}
\hline
\textbf{} & \textbf{Incompressible} & \textbf{Compressible} \\
\hline
PINC Transient & \cellcolor[HTML]{CAEDFB}65.9 ms ± 3.18 ms  & \cellcolor[HTML]{CAEDFB}504 ms ± 7.54 ms  \\
Num. Solution & \cellcolor[HTML]{CAEDFB}2.51 s ± 12.1 ms  & \cellcolor[HTML]{CAEDFB}1 min 35 s ± 3.68 s  \\
Time Ratio & \cellcolor[HTML]{CAEDFB}38.1 & \cellcolor[HTML]{CAEDFB}188.5 \\
\hline
Pressure & \cellcolor[HTML]{DAF2D0}95.68\% & \cellcolor[HTML]{DAF2D0}95.90\% \\
Velocity & \cellcolor[HTML]{DAF2D0}93.68\% & \cellcolor[HTML]{DAF2D0}93.92\% \\
\hline
\end{tabular}
\label{tab:transient_resume}
\end{table}

Finally, we emphasize a crucial aspect that supports the application of this methodology in monitoring, optimization, and control technologies: the PINC model is not only accurate in both steady-state and transient regimes but also computationally efficient. The inference time of the PINC model, compared to the numerical method, can be dozens or even hundreds of times faster.

This speedup is expected because the PINC requires only a single forward pass. Unlike numerical methods, which solve the full PDE system by sequentially updating variables at each time step, the PINC can perform all computations simultaneously. Once the inputs are defined—specifically the tuple \((\tilde{x}, \tilde{t}, \tilde{u}_0, \tilde{u})\), which represents the spatial position, time, initial condition, and control input, respectively—the PINC-Transient network processes the entire time evolution in a single forward pass. This means that, instead of iteratively marching through time as in traditional numerical schemes, the neural network directly outputs the full spatiotemporal solution in one batch. In the compressible transient case, which is the most computationally demanding, this capability results in a speedup ratio of 188.


%% file: 6-conclusion.tex
\section{Conclusion} \label{sec:conclusion}

This work presents an extension of the Physics-Informed Neural \eric{Nets for} Control (PINC) framework for modeling and controlling single-phase flow systems governed by PDEs. The approach follows a two-stage training methodology: a steady-state PINC model provides equilibrium solutions, which, \eric{in turn}, serve as \eric{target values for} the initial conditions \eric{during the training of} a transient PINC model. 
This strategy reduces the number of input features required for training, simplifying the neural network's modeling and parameterization. As a result, the method produces a highly accurate surrogate model that can be evaluated with a single forward pass, making it significantly faster than traditional numerical solvers and enabling its use in real-time applications such as monitoring, optimization, and control.

Numerical experiments confirm that the PINC model effectively captures both steady-state and transient behaviors of incompressible and compressible flows. The system can be simulated forward in time by cascading the transient PINC model across successive time windows, providing an efficient representation of flow dynamics. This forward simulation is then integrated into Model Predictive Control (MPC), leveraging the pre-trained transient PINC network as a predictive model for real-time flow optimization.

Future research directions include extending the PINC framework to multiphase flow systems and incorporating more complex fluid models and governing equations. 
\eric{Further applications of the proposed framework to other different systems described by PDEs will potentially show the generality of the approach.}
The findings of this study suggest that PINC is a viable approach for enhancing the efficiency and accuracy of flow control and monitoring strategies in engineering applications.

\section*{Acknowledgment}

This research was funded in part by Petróleo Brasileiro S.A.
(Petrobras) and Conselho Nacional de Desenvolvimento Científico e Tecnológico (CNPq) under grants 308624/2021-1 and 402099/2023-0.
